\documentclass[runningheads]{llncs}

\usepackage{eccv}

\usepackage{eccvabbrv}

\usepackage{graphicx}
\usepackage{booktabs}
\usepackage{multirow}
\usepackage{overpic}
\usepackage[accsupp]{axessibility}  

\usepackage[pagebackref,breaklinks,colorlinks,citecolor=eccvblue]{hyperref}

\usepackage{orcidlink}

\begin{document}

\title{Stable and Scalable Bundle Adjustment of Holistic 3D Structures} 

\titlerunning{Stable and Scalable Bundle Adjustment of Holistic 3D Structures}

\author{Shaohui Liu\inst{1} \and
R\'emi Pautrat\inst{2} \and
Daniel Barath\inst{1} \and
Richard Hartley\inst{3} \and
Viktor Larsson\inst{4} \and
Marc Pollefeys\inst{1,2}}

\authorrunning{S. Liu et al.}

\institute{\ \inst{1}ETH Zurich \quad \inst{2}Microsoft Spatial AI Lab \\ \inst{3}Australian National University \quad \inst{4}Lund University}

\maketitle

\begin{abstract}
  Bundle Adjustment (BA) is a cornerstone of 3D computer vision and has benefited from decades of advances in sparse optimization and numerical methods. It was originally developed for jointly optimizing camera intrinsics, poses and sparse 3D points. While extensions incorporate lines and other primitives, integrating richer geometric structures such as parallelism, coplanarity, or wireframes often introduces significantly increased computational cost and reduced numerical stability.
In this paper, we propose a unified framework that extends bundle adjustment to jointly optimize geometric features and higher-order relations. We first introduce a taxonomy that distinguishes scalable geometric \textit{features} with direct 2D measurements (e.g., points and lines), from \textit{groups} encoding higher-order relations (e.g., coplanarity, parallelism, etc.), where we show that groups can be modeled as camera-like entities within the bundle adjustment framework.
Building on this formulation, we propose that both group constraints and cross-feature relations (i.e., point–line associations) can be expressed through 2D reprojection measurements. By formulating group-induced and cross-feature reprojection errors, we preserve the sparsity structure of classical point-based BA under Schur elimination, while avoiding direct 3D regularization that degrades the conditioning and stability.
Experiments on both real-world and synthetic datasets demonstrate runtime performance comparable to classical point-only bundle adjustment, while producing significantly richer 3D structures and improved geometric accuracy. 

\end{abstract}

\section{Introduction}

\label{sec::intro}

\begin{figure}[t]
\centering
\setlength{\tabcolsep}{1pt}
\scriptsize

\begin{tabular}{cccccccc}
\includegraphics[width=0.12\linewidth]{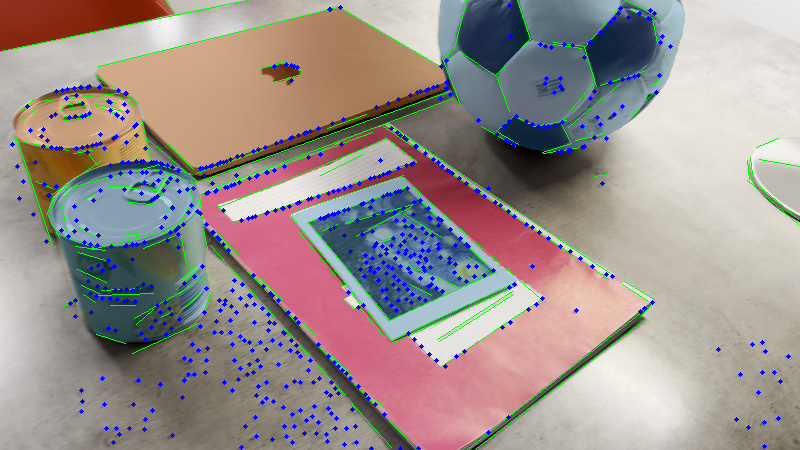} &
\includegraphics[width=0.12\linewidth]{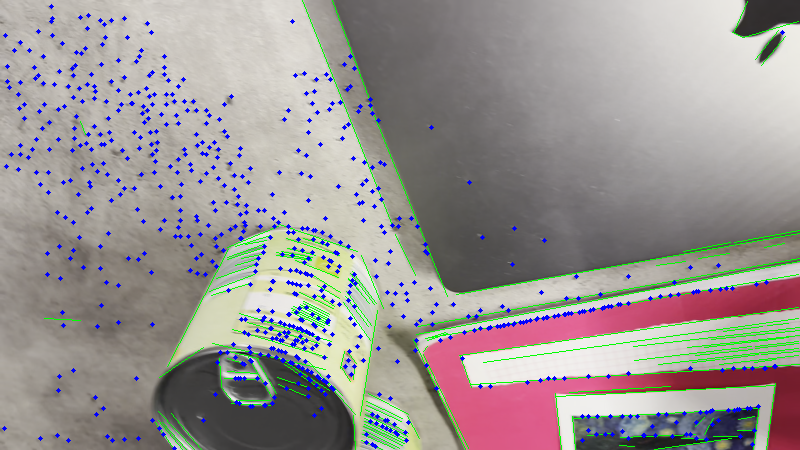} &
\includegraphics[width=0.12\linewidth]{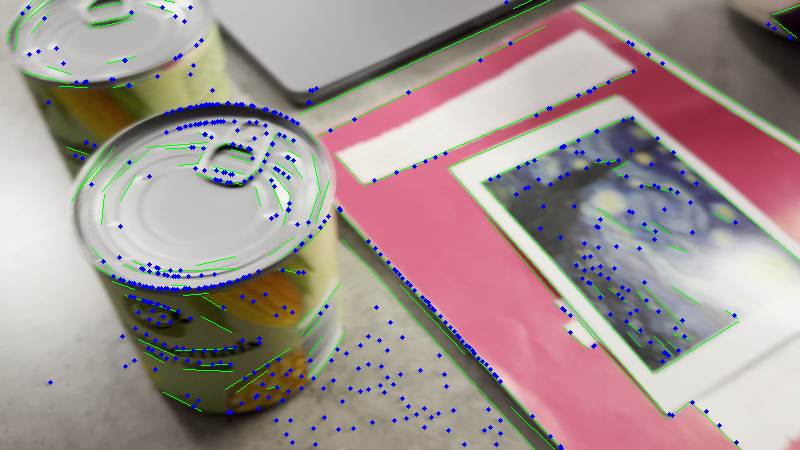} &
\includegraphics[width=0.12\linewidth]{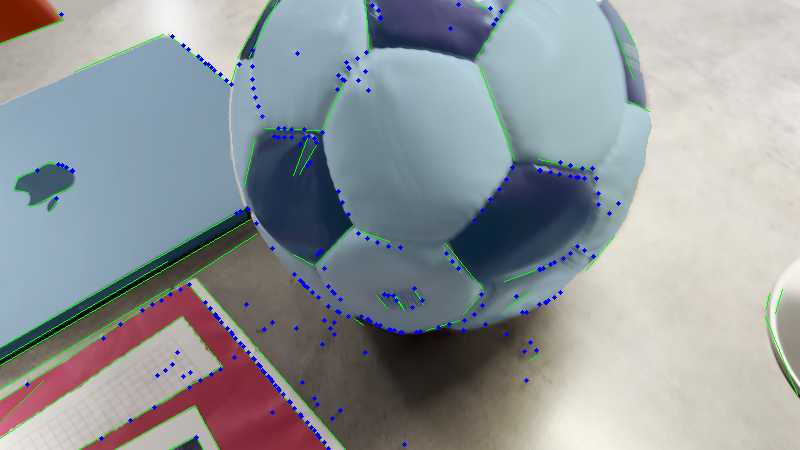} &
\includegraphics[width=0.12\linewidth]{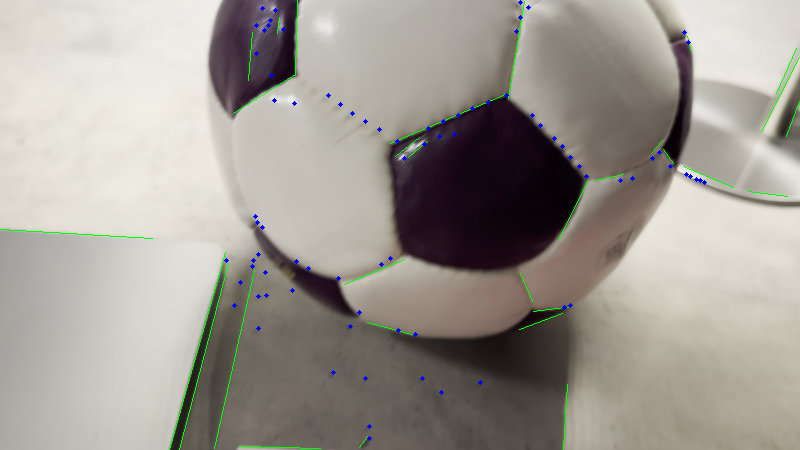} &
\includegraphics[width=0.12\linewidth]{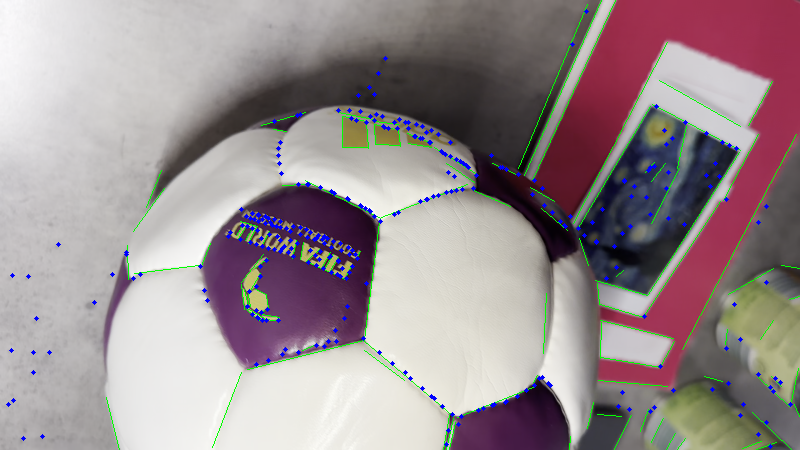} &
\includegraphics[width=0.12\linewidth]{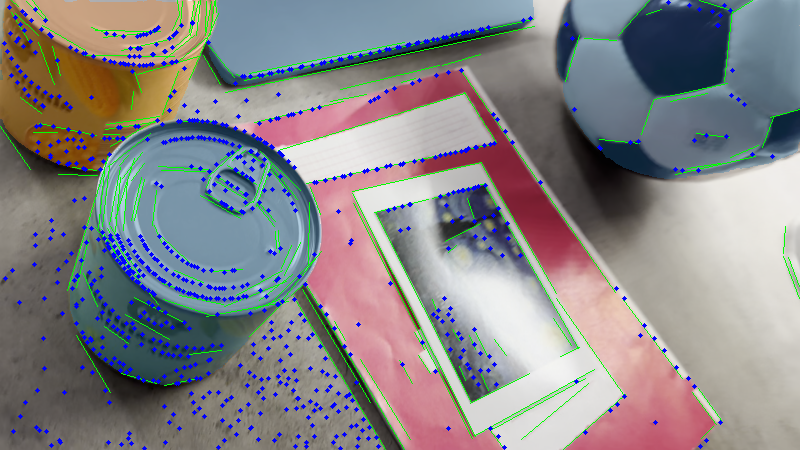} &
\includegraphics[width=0.12\linewidth]{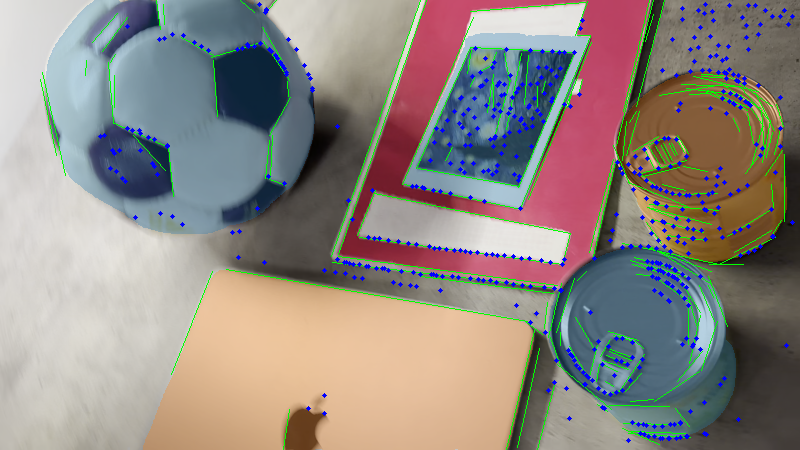}
\end{tabular}

\vspace{1pt}

\begin{tabular}{ccc}
\includegraphics[width=0.32\linewidth, trim={600 150 600 100}, clip]{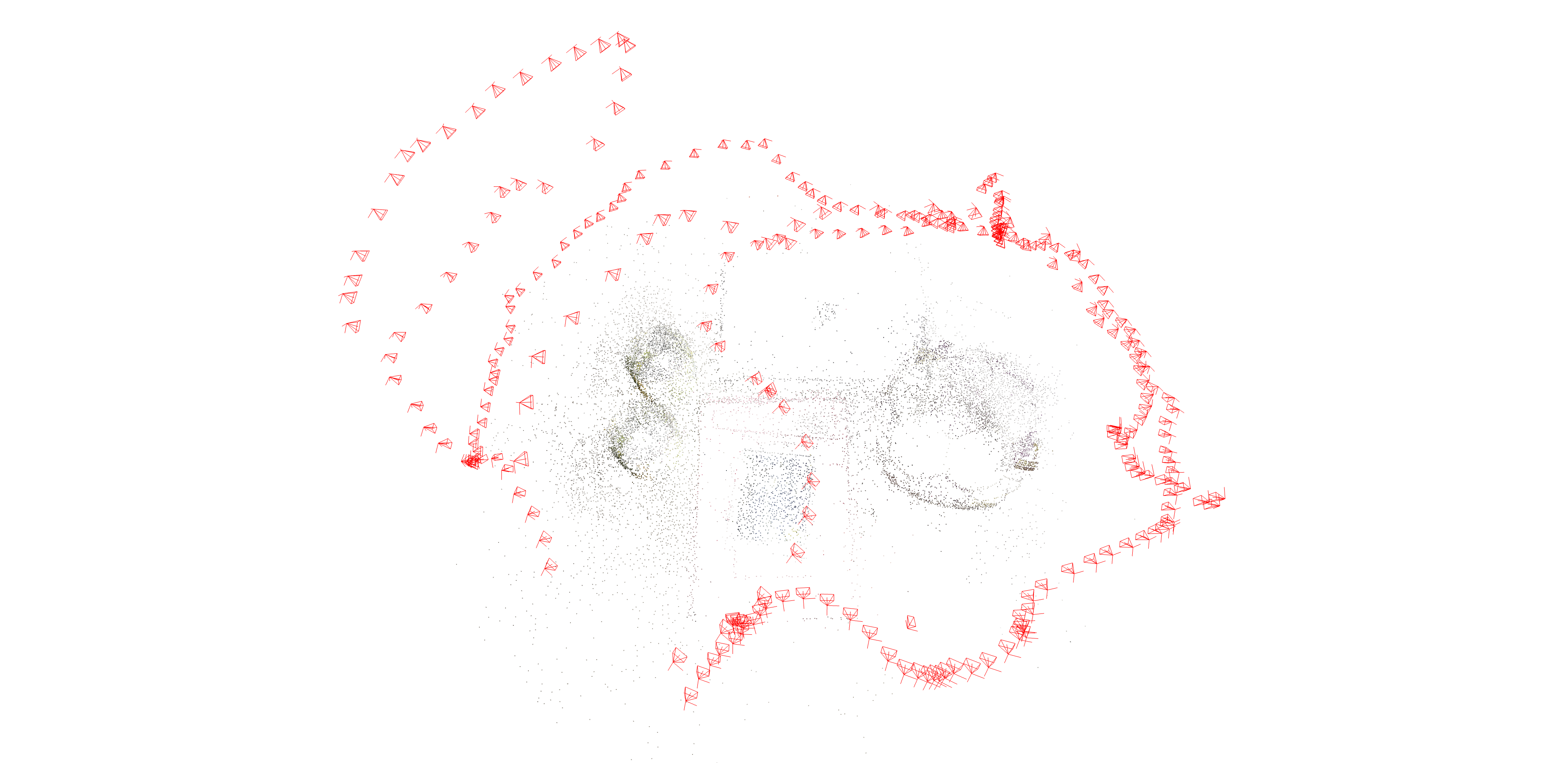} &
\includegraphics[width=0.32\linewidth, trim={600 150 600 100}, clip]{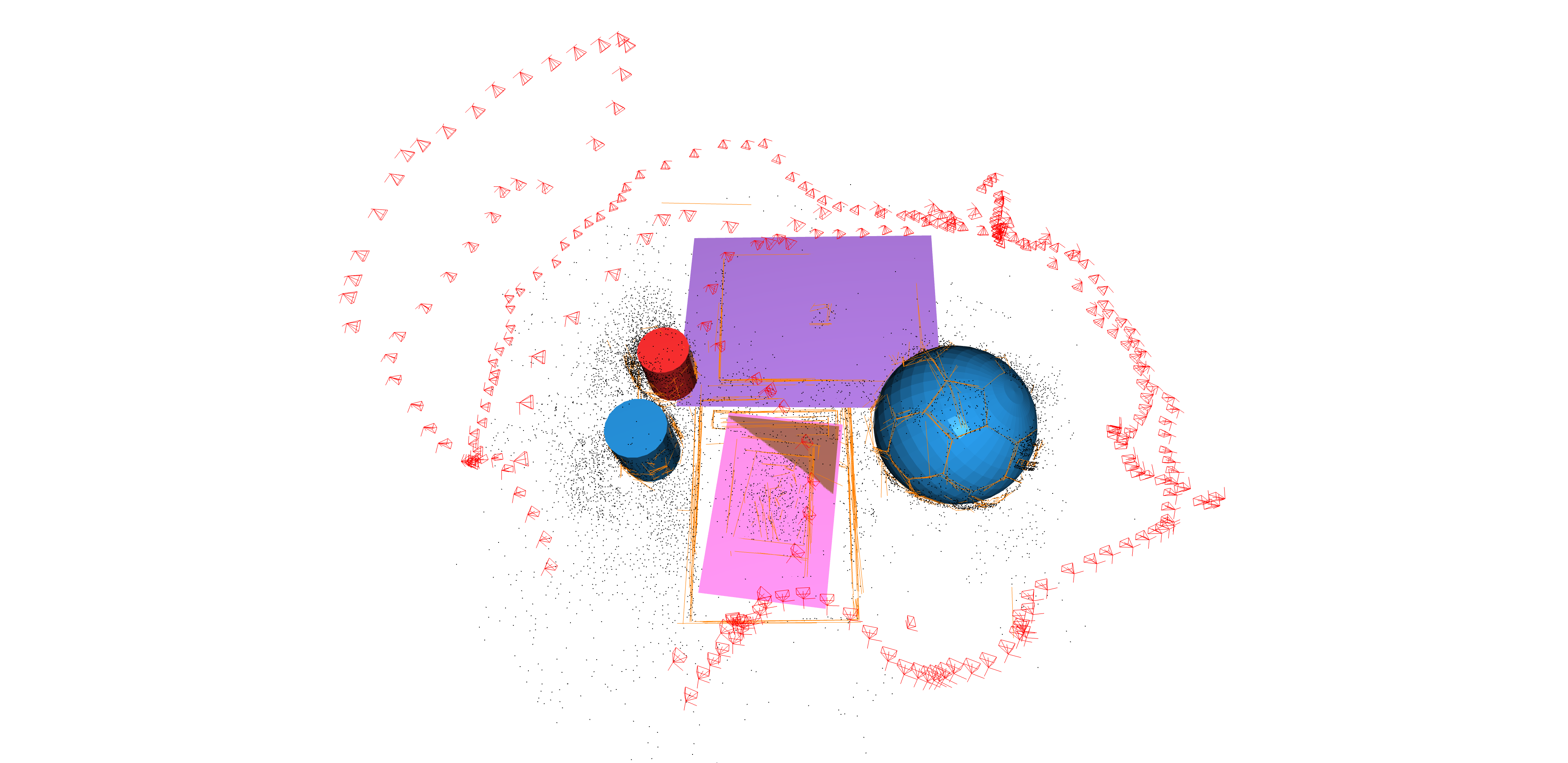} &
\includegraphics[width=0.32\linewidth, trim={600 150 600 100}, clip]{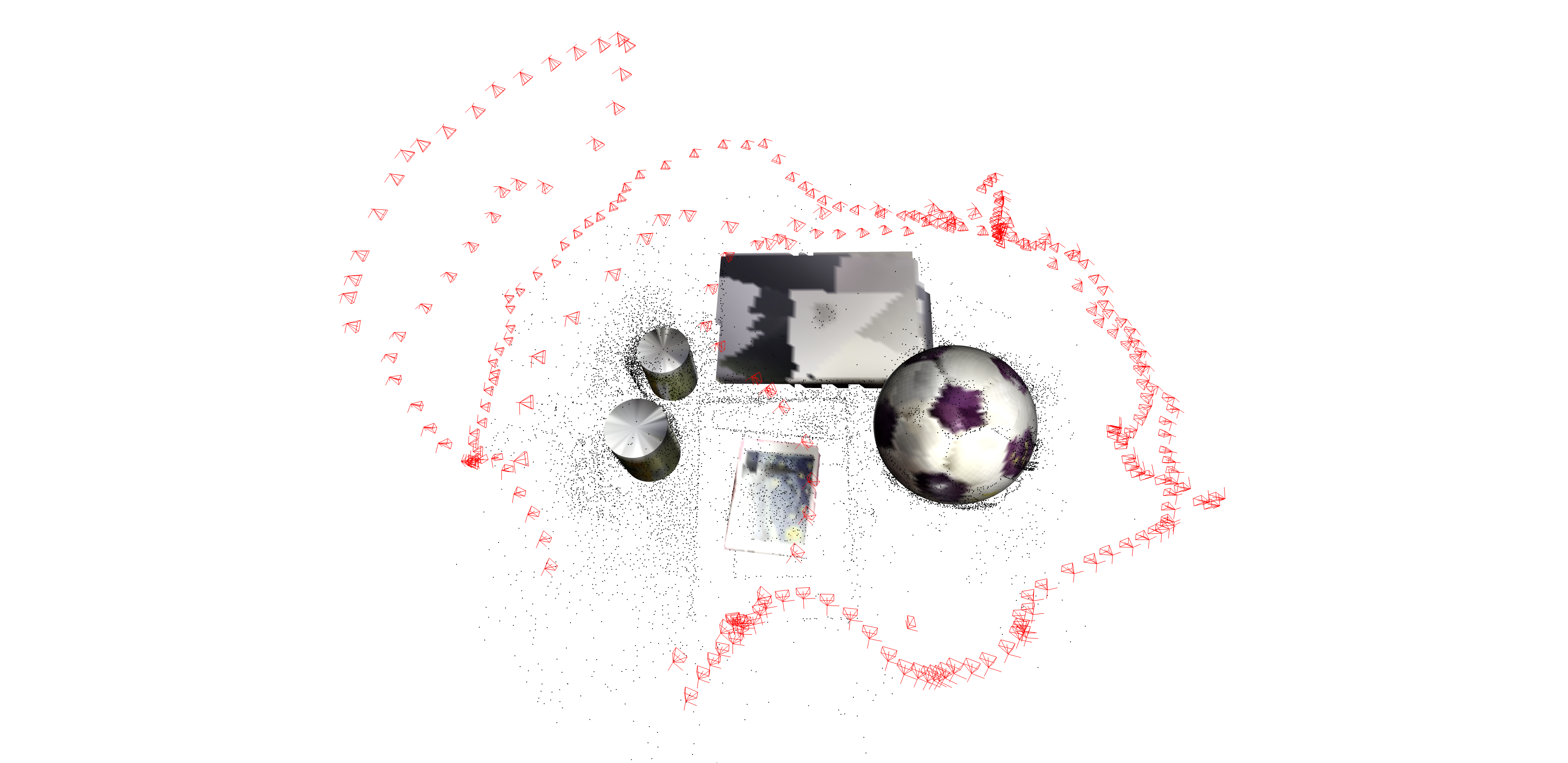} \\
Optimized 3D points & Optimized holistic 3D structures & Textured primitives
\end{tabular}

    \caption{\textbf{Bundle adjustment with structural constraints.} From multi-view images with detected points, lines, and groups (top), our proposed unified bundle adjustment framework is able to jointly optimize features (points/lines), camera poses, and structural constraints while preserving computational efficiency. This yields \textbf{richer optimizable sparse 3D reconstruction with geometric primitives} (planes, spheres, cylinders, etc.) beyond point clouds (left). 
    }
\label{fig::teaser}
\end{figure}

Bundle adjustment plays a central role in 3D reconstruction, jointly refining camera parameters and scene geometry by minimizing reprojection errors across all observations. Decades of research have produced highly efficient solvers that exploit the characteristic sparsity of the problem: because each 3D landmark couples only to its observing cameras, the Hessian over landmarks is block-diagonal, enabling Schur complement elimination that reduces the system to a compact camera-only problem. This structure has been the key to scaling bundle adjustment to large reconstructions with millions of points~\cite{agarwal2010bundle,wu2011visualsfm,schonberger2016structure}.

Yet real-world scenes, particularly man-made environments, contain geometric structures far richer than isolated point landmarks. Lines appear along edges and boundaries, often in parallel configurations governed by vanishing points~\cite{rother2002new}. Surfaces form planes and other primitives that constrain the positions of points and lines alike. At finer scales, points and lines meet at junctions to form wireframe structures~\cite{huang2018learning,zhou2019learning,xue2020holistically} that encode the geometric skeleton of buildings, furniture, and urban infrastructure. These structures provide powerful constraints that, if properly exploited, can improve reconstruction accuracy and completeness. However, incorporating them into bundle adjustment has remained a long-standing challenge due to two fundamental difficulties.

The first challenge is structural coupling. In classical bundle adjustment, the efficiency of the Schur complement relies on the fact that 3D landmarks are mutually independent: no residual couples two landmarks directly. Introducing geometric relations such as coplanarity, parallelism, orthogonality, or wireframe connectivity fundamentally violates this property. When two features share a geometric relation, their parameters become coupled, creating off-diagonal blocks in the landmark Hessian that destroy its block-diagonal structure. With many such associations, the efficient Schur elimination may break down entirely, leading to denser fill-in and increased computational cost. 

The second challenge is the scale ambiguity coming with 3D constraints. Even when sparsity is handled, formulating structural constraints as direct 3D penalties (\eg, point-to-plane distances~\cite{taguchi2013point,li2020leveraging}) mixes 3D metric residuals with 2D pixel-space terms, degrading conditioning due to scale mismatch and introducing ad hoc weighting. Because such residuals lack a principled noise model, there is no natural way to anchor structural entities through well-observed features. Moreover, 3D penalties break the scale gauge ambiguity of bundle adjustment, biasing the optimizer toward shrinking the reconstruction to reduce the penalty.

In this work, we present a unified bundle adjustment framework that integrates geometric structures while addressing both challenges. We introduce a taxonomy on features and groups, which distinguishes geometric features with direct 2D measurements (points, lines) from groups encoding higher-order relations (vanishing points, planes, conics, etc.). This taxonomy provides a formulation for incorporating diverse geometric entities under a single optimization scheme: virtual groups are added to the optimization problem and are placed alongside cameras in the non-eliminated block of the normal equations, absorbing the coupling between associated features and preserving the block-diagonal structure over the feature block.

Based on this formulation, we express both group and wireframe constraints as reprojection-based residuals in pixel space. Group-induced reprojection errors measure the pixel displacement from projecting a feature onto the group surface, while cross-feature reprojection errors use the observation of one feature to constrain another. This avoids direct 3D regularization and implicitly weights each constraint by the uncertainty (via Fisher information) of the involved features, naturally anchoring structural entities through more reliable associations.

We validate our framework through experiments on synthetic and real-world datasets. Our proposed framework is able to optimize on top of significantly richer reconstructions with holistic 3D structures (as in Fig. \ref{fig::teaser}), while achieving runtime performance comparable to classical bundle adjustment and consistent improvements on camera poses and geometry. Code for the proposed bundle adjustment framework and the resulting SfM system are made available at \href{https://github.com/cvg/limap}{\color{cyan}{https://github.com/cvg/limap}}, fully compatible with the COLMAP ecosystem.

\section{Related Work}
\noindent
\textbf{Advances in Bundle Adjustment.}
Bundle adjustment \cite{triggs2000bundle} has been widely studied and applied in 3D reconstruction for decades. Sparse matrix techniques and the Schur complement \cite{hartley2003multiple} have enabled efficient solutions by exploiting the bipartite structure between cameras and 3D points. 
There have been many efforts to improve its scalability via parallel/distributed computing \cite{agarwal2010bundle,wu2011multicore,fan2025daba} and advanced solvers \cite{lourakis2009sba,jeong2011pushing}. Other advances also include robust cost functions \cite{zach2014robust}, incremental formulation \cite{kaess2011isam2}, and differentiable BA for end-to-end learning \cite{tang2018ba,teed2021droid}. Building upon these advances, several open-sourced libraries have emerged and become standard tools, notably including g2o \cite{kummerle2011g}, Ceres solver \cite{ceres-solver}, GTSAM \cite{dellaert2012factor}, etc. These libraries work together with sparse linear algebra libraries \cite{chen2008algorithm, davis2019algorithm} and provide a nice ecosystem for modern developments of geometric 3D vision, enabling multiple reliable systems on structure-from-motion \cite{snavely2006photo,wu2011visualsfm,schonberger2016structure,moulon2016openmvg} and SLAM \cite{klein2007parallel,engel2014lsd,mur2015orb} across years. A separate line of work reformulates the linear system, either by marginalizing landmarks with
QR decomposition that never forms the Schur complement explicitly \cite{demmel2021square} or by expanding the matrix inverse as power series \cite{weber2023power}. Recently, bundle adjustment is also proved to be effective in refining the poses and geometry of feedforward 3D models \cite{wang2024vggsfm,wang2025vggt,chen2025back,wang2025pi,wang2026vggt}. In this work, we introduce a unified framework extending bundle adjustment to handle holistic 3D structures with higher-level primitives and relations.

~\\
\noindent
\textbf{Geometric Primitives in Optimization.}
While bundle adjustment is traditionally formulated over 3D points, extending it to general geometric primitives dates back to the 1990s \cite{hartley1996object}. However, most recent works that incorporate object-level primitives \cite{nicholson2018quadricslam,yang2019cubeslam,liao2020object,laidlow2022simultaneous,wu2023object} directly optimize high-dimensional variables representing the primitives through 2D modeling of primitive observations. Such formulations typically ignore the structural relationships between primitives and their supporting features, which prevents seamless integration with classical bundle adjustment over 3D points.

Among geometric primitives, 3D lines have been extensively studied. In \cite{bartoli2004framework}, Bartoli and Sturm formulate bundle adjustment over infinite 3D lines using an orthonormal parameterization of Plücker coordinates, which has since been adopted in many modern line-based optimization systems \cite{zuo2017robust,Liu_2023_LIMAP,liu2024robust,xu2025airslam}. These systems often further exploit structural cues such as vanishing points \cite{rother2002new,toldo2008robust}, Manhattan-world assumptions \cite{coughlan2000manhattan} and their variants \cite{schindler2004atlanta,li2023hong}, as well as learned monocular priors \cite{ke2026interacted}. In practice, these systems can scale to thousands of lines, which motivates us to treat lines as feature-level entities similar to points. 

Planes are another fundamental primitive, especially in man-made environments. Planar constraints have long been studied in geometric optimization, including formulations based on two-view homographies \cite{faugeras1988motion}, auto-calibration \cite{triggs1998autocalibration}, and explicit parameterization of 3D structures \cite{bartoli2003constrained}. The use of homography constraints has recently been revisited in \cite{zhou2023efficient} for computational efficiency. However, this formulation requires enumerating over image pairs. Explicit parameterization of 3D structures \cite{bartoli2003constrained} is conceptually appealing, but is sensitive to incorrect feature-plane associations, which are difficult to avoid in practice. Many recent works \cite{taguchi2013point,li2020leveraging,chen2023monocular,shu2023structure} therefore introduce soft point–plane regularization terms in 3D, which have also been extended to line–plane relationships \cite{li2020leveraging,bai2024consistent}. However, unlike point-to-plane formulations in LiDAR or RGB-D SLAM \cite{ma2016cpa,geneva2018lips,zhou2020fast,chen2022vip}, which benefit from principled noise models on 3D measurements, monocular systems typically measure errors in 3D space at absolute scale with heuristic weighting. In contrast, our framework jointly optimizes multiple types of primitives using a unified residual formulation in pixel space, which implicitly models the uncertainty of different feature–primitive associations.

~\\
\textbf{Modeling of Point-Line Associations.}
Wireframes are ubiquitous in man-made environments. The advent of deep neural networks has enabled reliable 2D wireframe prediction from single images \cite{zhou2019learning,xue2020holistically}. Wireframe representations have also proven beneficial for feature matching, either through junction-aware heuristics \cite{ma2019line,xu2025airslam} or learned approaches \cite{pautrat2023gluestick,ubingazhibov_2025}. Recent line-based reconstruction systems \cite{Liu_2023_LIMAP,liu2024robust} further incorporate point–line associations in optimization via 3D point-to-line distance regularization. However, these formulations introduce dense couplings between line segments that share junctions, disrupting the sparse structure and leading to expensive Schur complement operations, particularly with dense solvers. Our method addresses this limitation through a reformulation that preserves sparsity while fully exploiting wireframe constraints.

\section{A Unified Framework of Holistic Bundle Adjustment}

\subsection{Review of Classical Bundle Adjustment}
\label{sec::classical_ba}

Bundle adjustment (BA) jointly optimizes camera parameters and 3D structure by minimizing reprojection errors across all observations~\cite{triggs2000bundle}. Given $N$ images, each with pose $[R_i \mid \mathbf{t}_i]$ and associated (potentially shared) camera intrinsics $K_{c(i)}$, we denote the full camera parameters of image $i$ as $C_i = (K_{c(i)}, R_i, \mathbf{t}_i)$ and collect them in $\mathcal{C} = \{C_1, \ldots, C_N\}$. Given $M$ 3D points $\mathcal{X} = \{\mathbf{X}_1, \ldots, \mathbf{X}_M\}$, the classical BA objective is:
\begin{equation}
    E(\mathcal{C}, \mathcal{X}) = \sum_{(i,j) \in \mathcal{O}} \left\| \pi(C_i, \mathbf{X}_j) - \mathbf{x}_{ij} \right\|^2,
    \label{eq::classical_ba}
\end{equation}
where $\pi(C_i, \mathbf{X}_j)$ projects point $\mathbf{X}_j$ into camera $C_i$, $\mathbf{x}_{ij}$ is the corresponding 2D observation, and $\mathcal{O}$ is the set of all visibility pairs. Minimizing Eq.~(\ref{eq::classical_ba}) via the Levenberg-Marquardt algorithm leads to normal equations with a characteristic block structure:
\begin{equation}
    \begin{bmatrix} \mathbf{U} & \mathbf{W} \\ \mathbf{W}^\top & \mathbf{V} \end{bmatrix}
    \begin{bmatrix} \delta \mathcal{C} \\ \delta \mathcal{X} \end{bmatrix}
    = -\begin{bmatrix} \mathbf{g}_c \\ \mathbf{g}_x \end{bmatrix},
    \label{eq::normal_equations}
\end{equation}
where $\mathbf{U}$ and $\mathbf{V}$ are the camera-camera and point-point blocks of the approximate Hessian, $\mathbf{W}$ is the camera-point off-diagonal, and $\mathbf{g}_c$, $\mathbf{g}_x$ are the gradient vectors. A crucial property is that $\mathbf{V}$ is \textit{block-diagonal}: each 3D point appears only in residuals involving its observing cameras, creating no direct coupling between different points. This enables efficient Schur complement elimination, where points are marginalized to yield a reduced camera-only system:
\begin{equation}
    \left(\mathbf{U} - \mathbf{W}\mathbf{V}^{-1}\mathbf{W}^\top\right) \delta \mathcal{C} = -\mathbf{g}_c + \mathbf{W}\mathbf{V}^{-1}\mathbf{g}_x.
    \label{eq::schur}
\end{equation}
Since $\mathbf{V}^{-1}$ reduces to inverting small independent blocks, this elimination is both efficient and numerically stable. Modern solvers such as Ceres~\cite{ceres-solver} exploit this structure extensively. However, real-world scenes often exhibit richer geometric relations such as parallelism, coplanarity, and wireframe connectivity. We next introduce a taxonomy that distinguishes features from groups and show how both can be incorporated while preserving efficient Schur elimination. Figure \ref{fig::taxonomy} presents an illustration of the main concepts introduced in this paper.

\subsection{Features and Groups: A Taxonomy on Sparse 3D Structures}
\label{sec::taxonomy}

\begin{figure}[tb]
\centering
\includegraphics[trim={0 165 0 40}, clip, width=0.98\linewidth]{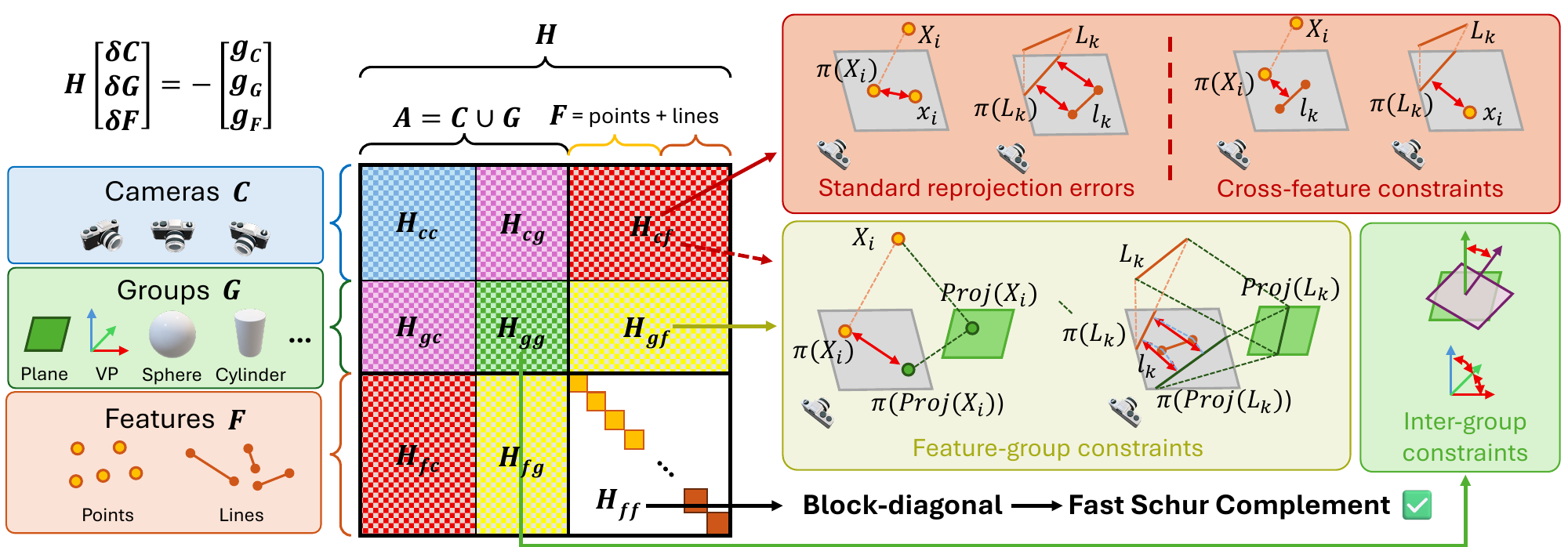}

\caption{\textbf{Taxonomy of features, groups, and cross-feature associations.} \textbf{Left:} Features $\mathcal{F}$ (points, lines) have direct 2D measurements; groups $\mathcal{G}$ (planes, VPs, 3D primitives) encode higher-order relations among features. \textbf{Middle:} approximate Hessian with block diagonal features $\mathbf{H}_{ff}$ matrix. \textbf{Right:} representations of the features, feature-group, cross-feature, and inter-group constraints in our optimization problem. }
\label{fig::taxonomy}
\end{figure}

We propose a taxonomy that categorizes geometric entities into \textit{features} and \textit{groups} based on their role in the normal equations of bundle adjustment. 

\noindent
\textbf{\textit{Features}} are geometric primitives with direct 2D measurements in each observing image. Points and lines are the canonical examples: a 3D point $\mathbf{X}_j$ is observed as a 2D keypoint $\mathbf{x}_{ij}$; a 3D line $\mathbf{L}_k$, parameterized in Pl\"ucker coordinates~\cite{bartoli2005structure}, is observed as a 2D line segment $\boldsymbol{\ell}_{ik}$. Adding lines to Eq.~(\ref{eq::classical_ba}) yields:
\begin{equation}
    E = \sum_{(i,j) \in \mathcal{O}_p} \left\| \pi_p(C_i, \mathbf{X}_j) - \mathbf{x}_{ij} \right\|^2 + \sum_{(i,k) \in \mathcal{O}_l} \left\| \rho\left(\pi_l(C_i, \mathbf{L}_k),\ \boldsymbol{\ell}_{ik}\right) \right\|^2,
    \label{eq::point_line_ba}
\end{equation}
where $\rho(\cdot, \cdot) \in \mathbb{R}^4$ computes the perpendicular distance vectors between the endpoints of the observed 2D segment and the projected 3D line. In the vanilla point-line bundle adjustment with reprojection errors, each feature contributes an independent diagonal block to $\mathbf{V}$ in the normal equation, since its residuals involve only itself and its observing cameras, without coupling the other features. 

\noindent
\textbf{\textit{Groups}} encode higher-order geometric relations among features. Vanishing points (VPs) encode parallelism among 3D lines, planes encode coplanarity of points and lines, and other primitives such as spheres, cylinders, cuboids, and object priors constrain features to lie on a parametric manifold. Unlike features, groups generally lack direct per-image 2D measurements in the classical sense. Instead, groups enforce consistency on its associated features: a VP group enforces directional consistency among its associated lines, while a plane group enforces coplanarity among its associated features. 

There are two common approaches to incorporating such group constraints into bundle adjustment. The first is explicit reparameterization \cite{bartoli2003constrained}, where features are constrained to lie exactly on a structure, \eg, reducing a 3D point on a plane from 3 to 2 degrees of freedom. However, this requires perfect feature-group associations prior to optimization, which is challenging in practice. The second introduces soft regularization terms \cite{taguchi2013point,li2020leveraging} that penalize deviations from structural constraints. This can be done either through direct pairwise or higher-order constraints between features (\eg, multiple points should be coplanar), or by introducing explicit group parameters with soft feature-group regularizations.

\noindent
\textbf{Computational Structure.} 
The two soft regularization approaches have different implications for  the structure of the normal equations. Direct constraints between features introduce off-diagonal blocks in the feature Hessian, breaking the block-diagonal property that enables efficient Schur elimination. For example, enforcing parallelism between two lines without a VP parameter creates a residual depending on both line parameters, filling in off-diagonal blocks and thus coupling them in the Hessian.

In contrast, introducing explicit group parameters avoids this coupling (\cref{fig::taxonomy}, middle). Each feature–group residual involves exactly one feature and one group, analogous to how a camera–feature residual involves one feature and one camera. The key is to place group parameters alongside cameras in the non-eliminated block. Given group parameters $\mathcal{G} = \{G_1, \ldots, G_P\}$, we form an augmented set $\mathcal{A} = \mathcal{C} \cup \mathcal{G}$ and rewrite Eq.~(\ref{eq::normal_equations}) as:
\begin{equation}
\begin{bmatrix} \mathbf{H}_{aa} & \mathbf{H}_{af} \\
\mathbf{H}_{fa} & \mathbf{H}_{ff} \end{bmatrix}
\begin{bmatrix} \delta \mathcal{A} \\ \delta \mathcal{F} \end{bmatrix}
= -\begin{bmatrix} \mathbf{g}_a \\ \mathbf{g}_f \end{bmatrix}.
\label{eq::augmented_normal}
\end{equation}
Since each residual involves at most one feature, $\mathbf{H}_{ff}$ remains block-diagonal. The Schur complement eliminates features to yield a reduced system over cameras and groups: $\mathbf{S} = \mathbf{H}_{aa} - \mathbf{H}_{af}\mathbf{H}_{ff}^{-1}\mathbf{H}_{fa}$.

Note that the categorization of entities into features and groups reflects computational considerations, not just geometric ones. A line could in principle be treated as a group enforcing collinearity among its associated points, but lines typically number in the thousands, and placing them in the non-eliminated block would make the reduced system $\mathbf{S}$ prohibitively large. Similarly, a VP could act as a feature if 2D direction measurements are available, and a plane normal could serve as a feature when estimated from monocular cues. However, such 2D measurements tend to be inaccurate, and groups like planes and VPs are typically few enough (\ie, ($|\mathcal{G}| \ll |\mathcal{F}|$)) to remain in the non-eliminated block without significant overhead. Nonetheless, when available, 2D measurements of groups are still compatible with our framework, as they contribute only to $\mathbf{H}_{aa}$ and do not affect the block-diagonal structure of $\mathbf{H}_{ff}$.

\noindent
\textbf{Inter-Group Relations.}
A further benefit of placing groups in $\mathcal{A}$ is that cross-group and camera-group constraints can be added at no cost to the Schur complement structure, since they involve only parameters within $\mathcal{A}$ and do not affect $\mathbf{H}_{ff}$. For instance, we employ orthogonality and parallelism constraints between VP pairs and between plane normal pairs to enforce Manhattan world priors, which contribute only to $\mathbf{H}_{aa}$, or more specifically $\mathbf{H}_{gg}$ in \cref{fig::taxonomy}.

\noindent
\textbf{Cross-Feature Associations (Wireframes).}
Beyond group-feature relations, features of different types can be directly associated with each other. In architectural and urban scenes, points frequently lie on lines at junctions, forming wireframe structures. These point-line associations encode geometric constraints that are complementary to group constraints: while groups relate features to a shared structural entity, cross-feature associations relate features directly to each other. Unlike group constraints, this coupling between features cannot be resolved through parameter ordering alone, as both entities belong to $\mathcal{F}$. We address this challenge in \cref{sec::cross_feature_ba}.

\subsection{Group Constraints in Bundle Adjustment}
\label{sec::group_ba}

Having established that groups can be placed alongside cameras to preserve the block-diagonal structure (\cref{sec::taxonomy}), we now design the group-feature residuals. For directional groups such as VPs, angular residuals between the line direction $\mathbf{d}_k$ (from the Pl\"ucker representation) and the VP direction $\mathbf{v}_g$ are scale-free and straightforward to integrate:
\begin{equation}
r_v = |\mathbf{d}_k \times \mathbf{v}_g|.
\label{eq::vp_angular}
\end{equation}

Positional groups such as planes are less straightforward. A natural formulation uses direct 3D residuals; for example, coplanarity can be enforced via the point-to-plane distance $r_\text{3D} = \mathbf{n}^\top \mathbf{X}_j - d$, where $\mathbf{n}$ and $d$ denote the plane normal and offset. However, such 3D constraints are problematic. Because reprojection errors are invariant to global scale, the optimizer can reduce 3D penalties by shrinking the reconstruction. In SfM, scale is typically fixed by constraining the translation of the initial image pair, which does not prevent the remaining structure from collapsing towards it. Moreover, balancing 3D and 2D terms depends on scene scale: proper weighting would require the 3D covariance of the features, which depends on viewing configurations and cannot be recovered from 2D observations alone. In contrast, 2D pixel noise is well characterized and approximately isotropic, making reprojection-based formulations easier to calibrate.

\noindent
\textbf{Group-Induced Reprojection Error.}
Motivated by the discussions above, we propose to formulate positional group constraints through pixel-space residuals that measure the reprojection difference caused by the group projection (\cref{fig::taxonomy}, right). For a plane group $G_g$ with normal $\mathbf{n}$ and offset $d$, we let the group-induced residual for an associated point $\mathbf{X}_j$ observed in camera $C_i$ be:
\begin{equation}
    r_g^{(p)} = \pi_p\!\left(C_i,\ \mathbf{X}_j\right) - \pi_p\!\left(C_i,\ \text{Proj}_{G_g}(\mathbf{X}_j)\right) \in \mathbb{R}^2,
    \label{eq::group_reproj_point}
\end{equation}
where $\text{Proj}_{G_g}(\mathbf{X}_j) = \mathbf{X}_j - \frac{\mathbf{n}^\top \mathbf{X}_j - d}{\|\mathbf{n}\|^2}\mathbf{n}$ projects the 3D point onto the plane surface. If the point lies on the plane, $\text{Proj}_{G_g}(\mathbf{X}_j) = \mathbf{X}_j$ and the residual vanishes. Otherwise, it measures the pixel displacement between the original and group-constrained reprojections. The same principle applies to associated lines: the residual is the difference between the standard line reprojection residual and that of the plane-projected line:
\begin{equation}
    r_g^{(l)} = \rho\!\left(\pi_l(C_i, \mathbf{L}_k),\ \boldsymbol{\ell}_{ik}\right) - \rho\!\left(\pi_l\!\left(C_i,\ \text{Proj}_{G_g}(\mathbf{L}_k)\right),\ \boldsymbol{\ell}_{ik}\right) \in \mathbb{R}^4.
    \label{eq::group_reproj_line}
\end{equation}

More generally, any geometric entity that defines a projection operator from 3D space onto a constraint surface can serve as a group in our framework, with the group-induced reprojection error following the same pattern. In all cases, each residual involves exactly one feature (in $\mathcal{F}$) and at most one group plus one camera (in $\mathcal{A}$), preserving the block-diagonal structure of $\mathbf{H}_{ff}$.

\noindent
\textbf{Statistical Interpretation: Group Anchoring through Implicit Weighting.}
Beyond avoiding scale ambiguity, the reprojection difference formulation provides a clean mechanism for anchoring the group surface through well-observed features. 
Concretely, let $\boldsymbol{\delta} = \mathbf{X}_j - \text{Proj}_{G_g}(\mathbf{X}_j) \in \mathbb{R}^3$ denote the displacement of a point from the group surface, and let $J_i = \partial \pi_p / \partial \mathbf{X} \big|_{C_i, \mathbf{X}_j} \in \mathbb{R}^{2 \times 3}$ be the projection Jacobian in view $i$. To first order, the group-induced residual (Eq.~(\ref{eq::group_reproj_point})) linearizes as:
\begin{equation}
    r_g^{(p)} \approx J_i \boldsymbol{\delta}.
    \label{eq::linearized_group}
\end{equation}
The effective cost of the group constraint, summed over all $N_j$ views observing the point, is therefore:
\begin{equation}
    E_g \approx \sum_{i} \| J_i \boldsymbol{\delta} \|^2 = \boldsymbol{\delta}^\top \!\left(\sum_{i} J_i^\top J_i \right) \boldsymbol{\delta}.
    \label{eq::implicit_mahalanobis}
\end{equation}
The matrix $\sum_i J_i^\top J_i$ is the Fisher information matrix of the point's 3D position, \ie, the inverse of its covariance $\boldsymbol{\Sigma}_{\mathbf{X}}^{-1}$ \cite{forstner2016photogrammetric}. Notably, this is the same matrix that arises from the standard reprojection residuals used to triangulate the point, so the group constraint inherits a geometrically consistent uncertainty model. The group-induced cost is thus equivalent to a Mahalanobis distance $\boldsymbol{\delta}^\top \boldsymbol{\Sigma}_{\mathbf{X}}^{-1} \boldsymbol{\delta}$. We want to note that under the common isotropic noise assumption, our group residual is equivalent to 3D point-to-plane distance with 3D point covariance computed from relinearized reprojection Jacobians at each iteration, which is closely related to the reduction perspective of implicit models in \cite{triggs2000bundle}. 

This directly determines how the group surface is estimated: well-observed features (with large Fisher information) produce large residuals when off-surface, anchoring the group to their positions, while poorly observed features contribute little regardless of displacement. The group parameters are therefore driven by reliable features, which in turn benefit from the structural consistency enforced by the group constraint. As shown in our experiments, this consistently improves both pose and geometry accuracy.

Note that our formulation cannot recover features with degenerate viewing geometry onto the group surface. A 3D formulation could push such features onto the surface, but since degenerate directions are precisely those where reprojection is insensitive, cameras do not benefit from this recovery.  Importantly, such features also pose minimal risk to group estimation: their weak Fisher information produces small residuals that have little influence on the group parameters.


\noindent
\textbf{Sparsity Analysis on $\mathbf{H}_{cg}$.}
While the group-induced reprojection residual involves camera, feature, and group parameters simultaneously, it does not worsen the sparsity pattern of the reduced system compared to a 3D formulation. A 3D point-to-plane residual $r = \mathbf{n}^\top \mathbf{X} - d$ has no direct dependence on camera parameters, contributing only to $\mathbf{H}_{gf}$ and $\mathbf{H}_{ff}$. However, after Schur complement elimination of features, indirect camera-group coupling arises in $\mathbf{H}_{cg}$ wherever a camera observes a feature associated with a group. Our 2D formulation creates this coupling directly, but the sparsity pattern is identical -- a nonzero entry in $\mathbf{H}_{cg}$ exists if and only if the camera sees some feature associated with the group. The structure of the reduced system $\mathbf{S}$ is thus unchanged.

\subsection{Cross-Feature Constraints in Bundle Adjustment}
\label{sec::cross_feature_ba}

As discussed in \cref{sec::taxonomy}, wireframe constraints couple features directly in $\mathbf{H}_{ff}$, breaking its block-diagonal structure. For example, enforcing a point-on-line constraint in 3D as in \cite{Liu_2023_LIMAP} introduces a residual that depends on both feature parameters, and the 3D point-to-line distance suffers from the same scale-dependent weight calibration issues as discussed in \cref{sec::group_ba}. We therefore propose a reformulation that preserves sparsity.

\noindent
\textbf{Cross-Feature Reprojection Error.}
Our key observation is that wireframe constraints can be reformulated as reprojection errors where only \textit{one} feature participates as an optimization variable, while the \textit{2D observation} of the other feature serves as a fixed measurement. Given a wireframe edge $(\mathbf{X}_j, \mathbf{L}_k)$, we define two complementary residuals across their respective visibility sets.

For an image $i$ where line $\mathbf{L}_k$ is observed as $\boldsymbol{\ell}_{ik}$ but point $\mathbf{X}_j$ is \textit{not} directly detected, the point-to-line residual measures the distance from the projected point to the observed line:
\begin{equation}
    r_{p \to l} = d\!\left(\pi_p(C_i, \mathbf{X}_j),\ \boldsymbol{\ell}_{ik}\right),
    \label{eq::point_to_line}
\end{equation}
where $d(\cdot, \cdot)$ is the perpendicular point-to-line distance in pixels. The optimization variables are $C_i$ (camera, in $\mathcal{A}$) and $\mathbf{X}_j$ (feature, in $\mathcal{F}$); the 2D line observation $\boldsymbol{\ell}_{ik}$ enters as a constant measurement. Intuitively, if the point lies on the line, it should project near the observed line in any view where the line is visible, even in views where the point itself is not detected. As with the group-induced reprojection error (Sec.~\ref{sec::group_ba}), this 2D formulation avoids scale-dependent 3D distances and implicitly weights the constraint by observation geometry, with well-observed features providing stronger anchoring signals (Eq.~(\ref{eq::implicit_mahalanobis})).

Conversely, for an image $i$ where point $\mathbf{X}_j$ is observed as $\mathbf{x}_{ij}$ but line $\mathbf{L}_k$ is not detected, the line-to-point residual is:
\begin{equation}
    r_{l \to p} = d\!\left(\mathbf{x}_{ij},\ \pi_l(C_i, \mathbf{L}_k)\right),
    \label{eq::line_to_point}
\end{equation}
where $\mathbf{x}_{ij}$ is a constant measurement and $\mathbf{L}_k$ is the optimization variable. In images where both features are directly observed, we omit the cross-feature terms as their standard reprojection residuals (Eq.~(\ref{eq::point_line_ba})) already provide the constraint. Note that this cross-feature reprojection error formulation inherit the same statistical interpretation of the classical reprojection error and thus naturally support more advanced noise models.

\noindent
\textbf{Sparsity Analysis.}
Each cross-feature residual involves exactly one feature and one camera. The Jacobian of $r_{p \to l}$ has nonzero blocks only for $C_i$ and $\mathbf{X}_j$; that of $r_{l \to p}$ only for $C_i$ and $\mathbf{L}_k$. No off-diagonal blocks arise between point $j$ and line $k$ in $\mathbf{H}_{ff}$, preserving its block-diagonal structure. The constraint is mediated through the 2D observation of the partner feature, decoupling the features while still enforcing their geometric relation. Note that cross-feature constraints extend the effective visibility graph, making the reduced system $\mathbf{S}$ denser in camera–camera blocks, but efficient Schur elimination remains preserved.

\subsection{Discussions and Applications}
\label{sec::discussions}

\noindent
\textbf{Convergence and Termination.}
A practical benefit of formulating constraints in pixel space is simplified convergence monitoring. When mixing 3D regularization terms (in metric units) with 2D reprojection errors (in pixels), the residuals have different magnitudes and convergence dynamics, complicating cost-based termination criteria such as the relative cost change $|\Delta E| / E$. With our unified pixel-space formulation, the total cost has a consistent interpretation and such criteria apply directly. For the remaining angular residuals (\eg, VP constraints in Eq.~(\ref{eq::vp_angular}) and cross-group angular constraints), we isolate their contribution and monitor convergence based on the reprojection terms only. 


\noindent
\textbf{Structure-from-Motion Pipeline.}
Our bundle adjustment framework integrates naturally into incremental structure-from-motion. After registering each new image and triangulating features and groups, the system performs local bundle adjustment with the full set of constraints, including features, groups, and wireframe edges. Image selection leverages both point and line visibility, and camera registration uses both point and line correspondences. This yields an end-to-end SfM pipeline where geometric structures participate from the earliest stages of reconstruction. 
We also want to note that, the sparsity analysis and Schur elimination strategy extend directly to global positioning \cite{zhuang2018baseline,pan2024global} in global SfM, as cameras, features, and groups share the same computational structure.

\noindent
\textbf{Discussion: Robustness to Noisy Upstream Associations.}
Our work focuses on optimization once structural prios are available, rather than solving the problem of adaptive model selection. We therefore employ an structure association strategy that favors precision over recall and optimize these residuals with robust losses. Importantly, robust thresholding becomes easier in our pixel-level formulation than in mixed 2D/3D formulations, especially for SfM with its scale ambiguity. Nonetheless, the reliance on reasonably good quality of associations is an important practical limitation of our framework. We also want to note the distinction between measurement noise and association reliability, where the former is handled in a principled way through how our residuals are formulated, while the latter is closely related to model selection and benefits from future advances in estimating confidences of the extracted structural constraints.

\section{Experiments}

\noindent
\textbf{Implementation Details.}
Our framework is agnostic to the choice of feature and group detection and matching methods. We use ALIKED~\cite{zhao2023aliked} / DeepLSD~ \cite{Pautrat_2023_DeepLSD} and LightGlue~\cite{lindenberger2023lightglue} / GlueStick~\cite{pautrat2023gluestick} for point / line detection and matching. Vanishing points are detected with JLinkage \cite{toldo2008robust}, and planes with a custom segmentation method fitting planes to predicted depth and surface normals of MoGe-2~\cite{wang2025moge,wang2025moge2}. Images are resized so that the longer side is at most 800 pixels, to satisfy model constraints.
Within each image, feature-group associations are obtained during detection (e.g., lines assigned to VPs, points and lines to plane segments), while point-line associations are derived from spatial incidences. Across images, group matchings are established either by voting over matched features or by IoU checks of warped segmentation masks via dense matching \cite{edstedt2024roma,zhang2025ufm}.
To construct 3D constraints from 2D observations, we aggregate feature-group and cross-feature associations across images through voting following \cite{Liu_2023_LIMAP}. 
Group constraints count images where a feature observation is associated with a group, while wireframe constraints count images where a point-line pair forms a junction. Constraints are instantiated when the number of such images exceeds 3, and susbsequently optimzied with a robust Cauchy loss. 
Please refer to more details in our supplementary material. 

Our proposed BA is built on top of Ceres solver \cite{ceres-solver} with analytical Jacobians. The SfM pipeline follows COLMAP \cite{schonberger2016structure}, which automatically selects the numerical solver depending on number of images. Runtime experiments are conducted on a workstation with an 8-core Intel Core i7-10700K CPU at 3.80 GHz.

\begin{figure}[tb]
\centering
\begin{tabular}{ccc}
\includegraphics[width=0.32\columnwidth]{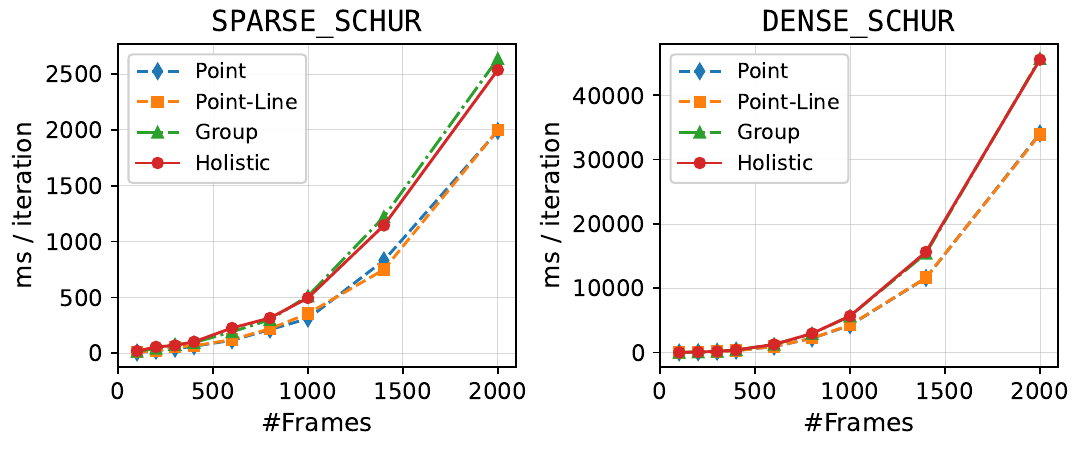} &
\includegraphics[width=0.32\columnwidth]{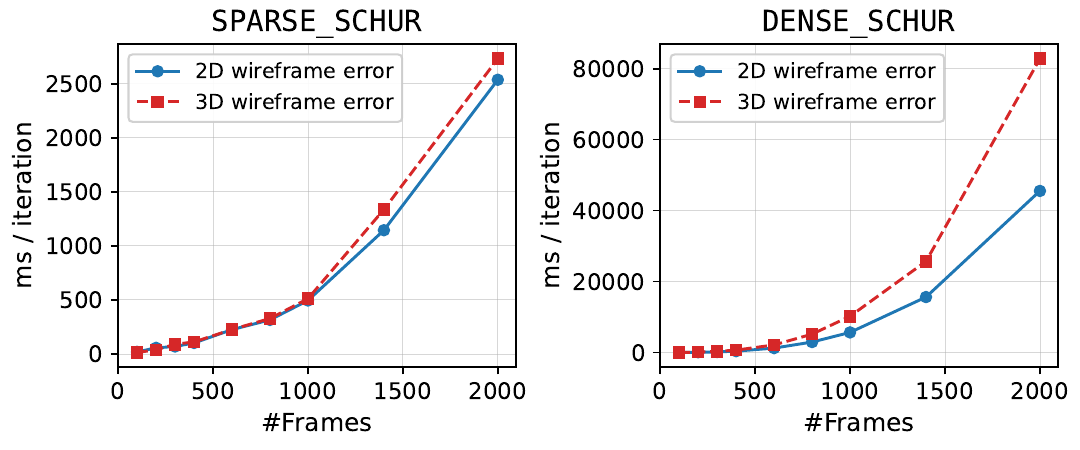} &
\includegraphics[width=0.32\columnwidth]{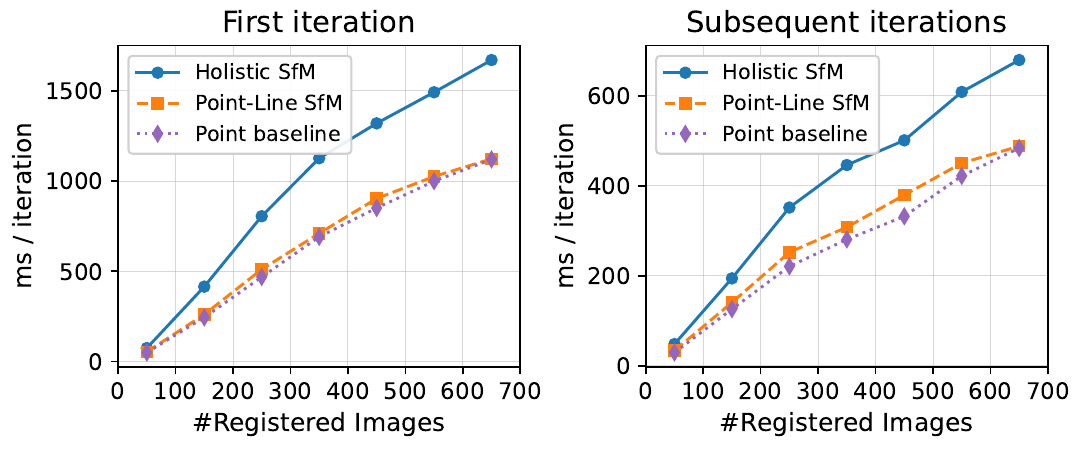} \\
(a) Scaling w.r.t. \#images & (b) 2D vs 3D wireframe cost & (c) Analysis on 1DSfM~\cite{wilson2014robust} \\
\end{tabular}
\caption{\textbf{Bundle adjustment runtime analysis.} (a) Per-iteration cost scaling with number of images. (b) Comparison between 2D vs 3D wireframe cost. (c) First iteration vs. subsequent iterations of BA on 1DSfM, averaged across 3 scenes. }
\label{fig:ba_benchmark}
\end{figure}

\begin{table}[tb]
\centering
\scriptsize
\setlength{\tabcolsep}{4pt}
\caption{\textbf{Runtime and reconstruction statistics on 1DSfM}~\cite{wilson2014robust}.
The \textit{Point baseline} is equivalent to COLMAP with only one change: replacing the gradient-norm termination criterion with the relative cost change based criterion. The two other methods use the same criterion as the point baseline. We report runtime, BA statistics, reconstruction size (\#Reg: number of registrations, \#pts: 3D points, \#lns: lines, \#grps: group primitives), and mean reprojection error over 3D point observations (in pixels).}
\resizebox{\columnwidth}{!}{%
\begin{tabular}{ll rrrr c r@{/}l rr r}
\toprule
& & \multicolumn{4}{c}{Runtime \& Convergence} & \multicolumn{6}{c}{Reconstruction statistics} \\
\cmidrule(lr){3-6} \cmidrule(lr){7-12}
Scene & Method & SfM (min) & ms/iter & iter/BA & \#BA & \#Reg & \multicolumn{2}{c}{\#pts/\#obs} & \#lns & \#grps & err (px) \\
\midrule
\multirow{4}{*}{Tower of London}
& COLMAP reference & 63.6 & — & 16.2 & 194 & 935 / 1322 & 97k & 731k & — & — & 1.08 \\
& Point baseline & 46.4 & 132.5 & 4.8 & 196 & 909 / 1322 & 99k & 731k & — & — & 1.04 \\
& Point-Line SfM & 54.6 & 158.9 & 4.8 & 205 & 927 / 1322 & 99k & 732k & 2.9k & — & 1.02 \\
& Holistic SfM (Ours) & 59.2 & 214.0 & 11.8 & 226 & 963 / 1322 & 99k & 731k & 2.8k & 117 & 1.03 \\
\midrule
\multirow{4}{*}{Madrid Metropolis}
& COLMAP reference & 45.3 & — & 21.7 & 225 & 742 / 1061 & 59k & 462k & — & — & 1.13 \\
& Point baseline & 31.5 & 136.3 & 5.4 & 233 & 752 / 1061 & 65k & 476k & — & — & 1.07 \\
& Point-Line SfM & 34.8 & 153.5 & 5.6 & 249 & 753 / 1061 & 63k & 465k & 2.1k & — & 1.05 \\
& Holistic SfM (Ours) & 37.6 & 192.1 & 9.5 & 253 & 750 / 1061 & 65k & 469k & 2.1k & 42 & 1.06 \\
\midrule
\multirow{4}{*}{Gendarmenmarkt}
& COLMAP reference & 68.8 & — & 14.6 & 234 & 943 / 1173 & 85k & 846k & — & — & 1.16 \\
& Point baseline & 43.5 & 281.5 & 4.2 & 248 & 950 / 1173 & 89k & 847k & — & — & 1.10 \\
& Point-Line SfM & 45.7 & 274.7 & 4.5 & 266 & 947 / 1173 & 90k & 851k & 3.8k & — & 1.10 \\
& Holistic SfM (Ours) & 63.5 & 394.3 & 11.3 & 281 & 951 / 1173 & 88k & 846k & 3.7k & 91 & 1.10 \\
\bottomrule
\end{tabular}%
}
\label{tab:benchmark_runtime}
\end{table}

\subsection{Benchmarking Holistic Bundle Adjustment}
\noindent
\textbf{Synthetic Benchmark.}
To analyze scaling behavior in isolation, we construct synthetic BA problems by varying the number of images from 100 to 2000. For each configuration, we generate points, lines, planes, and wireframe edges with counts scaling proportionally. Each feature is observed with an average track length of 4-8 images. Please refer to details in supplementary materials.

We compare four BA configurations of increasing complexity: \textit{Point} (point-only), \textit{Point-Line} (points and lines), \textit{Groups} (adding group constraints from \cref{sec::group_ba}), and \textit{Holistic} (further adding
  wireframe constraints from \cref{sec::cross_feature_ba}).
\Cref{fig:ba_benchmark}(a) shows per-iteration cost as a function of problem sizes. All bundle adjustment configurations scale similarly under both \texttt{SPARSE\_SCHUR} ($\sim n^{1.6-1.9}$) and \texttt{DENSE\_SCHUR} ($\sim n^{2.6-2.8}$), indicating that our formulation preserves the asymptotic complexity of classical BA. With \texttt{DENSE\_SCHUR}, runtime is strictly determined by the size of the reduced system, with \textit{Point} and \textit{Point-Line}, \textit{Groups} and \textit{Holistic} having nearly identical cost. This supports our analysis that groups behave similarly to camera variables in the normal equations.
\Cref{fig:ba_benchmark}(b) compares the per-iteration runtime of our 2D reprojection-based wireframe formulation with a 3D endpoint-distance formulation. With \texttt{SPARSE\_SCHUR}, the 2D formulation is moderately faster as it avoids direct coupling between features. With \texttt{DENSE\_SCHUR}, the gap widens because the 3D formulation produces significantly denser fill-in in the reduced system.

\begin{table}[tb]
\centering
\scriptsize
\setlength{\tabcolsep}{3pt}
\caption{\textbf{Geometry evaluation on Hypersim~\cite{roberts:2021} (8 scenes).} Points and lines are measured against ground-truth meshes. }
\resizebox{\columnwidth}{!}{%
\begin{tabular}{l cccc ccc ccc}
\toprule
& \multicolumn{4}{c}{Points} & \multicolumn{6}{c}{Lines} \\
\cmidrule(lr){2-5} \cmidrule(lr){6-11}
& \multicolumn{3}{c}{Inlier (\%) $\uparrow$} & Med. $\downarrow$ & \multicolumn{3}{c}{Length Recall (m) $\uparrow$} & \multicolumn{3}{c}{Precision (\%) $\uparrow$} \\
\cmidrule(lr){2-4} \cmidrule(lr){5-5} \cmidrule(lr){6-8} \cmidrule(lr){9-11}
& @1mm & @5mm & @10mm & (mm) & @1mm & @5mm & @10mm & @1mm & @5mm & @10mm \\
\midrule
Per-point/per-line refinement & 9.3 & 31.7 & 45.6 & 15.10 & 72.1 & 222.5 & 301.2 & 21.3 & 59.4 & 77.3 \\
+ Group constraints & 11.5 & 35.6 & 48.5 & 13.80 & 93.2 & 248.8 & 315.5 & 27.3 & \textbf{65.5} & \textbf{81.0} \\
+ Group and wireframe constraints & \textbf{11.9} & \textbf{36.3} & \textbf{48.9} & \textbf{13.55} & \textbf{93.3} & \textbf{249.5} & \textbf{315.7} & \textbf{27.3} & \textbf{65.5} & \textbf{81.0} \\
\bottomrule
\end{tabular}%
}
\label{tab:geometry_hypersim}
\end{table}

\begin{table}[tb]
\centering
\scriptsize
\caption{\textbf{Point reconstruction quality on ETH3D\cite{schops2017multi} (13 scenes).} Reported against laser scans using the official ETH3D multi-view evaluation benchmark.}
\resizebox{\columnwidth}{!}{%
\begin{tabular}{l cccc cccc cccc}
\toprule
& \multicolumn{4}{c}{Accuracy (\%) $\uparrow$} & \multicolumn{4}{c}{Completeness (\%) $\uparrow$} & \multicolumn{4}{c}{F1 (\%) $\uparrow$} \\
\cmidrule(lr){2-5} \cmidrule(lr){6-9} \cmidrule(lr){10-13}
& @1cm & @2cm & @5cm & @10cm & @1cm & @2cm & @5cm & @10cm & @1cm & @2cm & @5cm & @10cm \\
\midrule
Per-point refinement & 45.17 & 58.70 & 75.53 & 84.57 & 0.14 & 0.74 & 4.10 & 11.09 & 0.29 & 1.46 & 7.66 & 19.25 \\
+ Group and wireframe constraints & \textbf{47.05} & \textbf{60.59} & \textbf{76.84} & \textbf{85.40} & \textbf{0.15} & \textbf{0.76} & \textbf{4.18} & \textbf{11.21} & \textbf{0.30} & \textbf{1.50} & \textbf{7.81} & \textbf{19.47} \\
\bottomrule
\end{tabular}%
}
\label{tab:geometry_eth3d}
\end{table}

\noindent
\textbf{Real-World Benchmark.}
We further evaluate three large-scale scenes from 1DSfM~\cite{wilson2014robust} within a full incremental SfM pipeline (\cref{tab:benchmark_runtime}). As discussed in \cref{sec::discussions}, we use the relative cost change as the BA termination criterion, which empiricially stops earlier than COLMAP, whose termination relies on the gradient norm. Our method registers a comparable or slightly larger number of images while maintaining similar reprojection errors. The higher iteration count per BA call (e.g., 11.8 vs.\ 4.8 on Tower of London) reflects the richer constraint set where structural constraints generally require more iterations for convergence, but the overall pipeline runtime remains $\sim$ 1.3$\times$ of the point baseline, substantially lower than the 2--4$\times$ overhead over COLMAP reported in \cite{liu2024robust}. \Cref{fig:ba_benchmark}(c) shows the BA cost breakdown on 1DSfM. The first iteration of each BA call performs symbolic analysis (sparsity pattern computation and ordering), while subsequent iterations require only numeric factorization. This symbolic overhead is amortized across iterations, after which the steady-state numeric cost dominates.

\subsection{Applications}

\begin{table}[tb]
\centering
\scriptsize
\setlength{\tabcolsep}{3pt}
\caption{\textbf{Evaluating the effectiveness of bundle adjustment in full structure-from-motion pipelines}. Camera pose accuracy is measured by relative pose AUCs.}
\resizebox{\columnwidth}{!}{%
\begin{tabular}{l ccc ccc}
\toprule
& \multicolumn{3}{c}{Hypersim (8 scenes)} & \multicolumn{3}{c}{ScanNet++ (20 scenes)} \\
\cmidrule(lr){2-4} \cmidrule(lr){5-7}
& AUC@3\textdegree & AUC@5\textdegree & AUC@10\textdegree & AUC@3\textdegree & AUC@5\textdegree & AUC@10\textdegree \\
\midrule
COLMAP & 89.3 & 90.4 & 92.2 & 84.0 & 86.0 & 87.5 \\
Point-Line SfM & 92.2 & 93.2 & 94.0 & 84.7 & 86.7 & 89.2 \\
Holistic SfM (Ours) & \textbf{92.5} & \textbf{93.4} & \textbf{94.2} & \textbf{87.4} & \textbf{89.4} & \textbf{90.6} \\
\midrule
& \multicolumn{3}{c}{ETH3D (11 scenes)} & \multicolumn{3}{c}{7Scenes (7 scenes)} \\
\cmidrule(lr){2-4} \cmidrule(lr){5-7} & AUC@3\textdegree & AUC@5\textdegree & AUC@10\textdegree & AUC@3\textdegree & AUC@5\textdegree & AUC@10\textdegree \\
\midrule
COLMAP & 50.2 & 66.4 & 77.6 & 22.9 & 48.7 & 75.5 \\
Point-Line SfM & 49.4 & 67.1 & 79.6 & 23.1 & 48.9 & 76.8 \\
Holistic SfM (Ours) & \textbf{50.4} & \textbf{68.1} & \textbf{80.4} & \textbf{24.3} & \textbf{50.8} & \textbf{78.4} \\
\bottomrule
\end{tabular}%
}
\label{tab:sfm_results}
\end{table}

\begin{table}[tb]
\centering
\scriptsize
\setlength{\tabcolsep}{3pt}
\caption{\textbf{Ablation of the group-induced 2D reprojection error vs direct 3D constraints} via geometry evaluation on ETH3D and SfM evaluation on ScanNet++.}
\begin{tabular}{l ccc cccc}
\toprule
& \multicolumn{3}{c}{ETH3D Acc.\ (\%) $\uparrow$} & \multicolumn{4}{c}{ScanNet++ AUC $\uparrow$} \\
\cmidrule(lr){2-4} \cmidrule(lr){5-8}
Method & @1cm & @5cm & @10cm & @1\textdegree & @3\textdegree & @5\textdegree & @10\textdegree \\
\midrule
Point BA & 45.17 & 75.53 & 84.57 & 66.1 & 84.0 & 86.0 & 87.5 \\
\midrule
Holistic BA w/. 3D group error ($w{=}1$) & 45.17 & 75.55 & 84.58 & 67.4 & 85.0 & 87.1 & 89.1 \\
Holistic BA w/. 3D group error ($w{=}10$) & 45.24 & 75.61 & 84.63 & 68.1 & 85.0 & 89.0 & 90.5 \\
Holistic BA w/. 3D group error ($w{=}100$) & 45.72 & 76.12 & 84.88 & \underline{68.3} & \textbf{87.8} & \textbf{89.9} & \textbf{91.1} \\
Holistic BA w/. 3D group error ($w{=}1000$) & \underline{46.69} & \underline{76.71} & \underline{85.15} & 67.1 & 85.1 & 87.1 & 88.3 \\
\midrule
Holistic BA (Ours) & \textbf{47.05} & \textbf{76.84} & \textbf{85.40} & \textbf{69.4} & \underline{87.4} & \underline{89.4} & \underline{90.6} \\
\bottomrule
\end{tabular}
\label{tab:ablation_group_3dcost}
\end{table}

\noindent
\textbf{Geometry Evaluation.}
To study the benefits of group and wireframe constraints on the resulting 3D maps, we evaluate reconstruction quality against ground-truth geometry on the first eight scenes of Hypersim \cite{roberts:2021}, following the protocol of \cite{liu2024robust}. Points and lines are measured against ground-truth meshes. For points, we report the inlier ratio (percentage within threshold) and median distance. For lines, length recall is the average correct length in meters across scenes, while precision is the fraction of total reconstructed length within threshold. The results, reported in \cref{tab:geometry_hypersim}, show that group constraints consistently improve the accuracy of reconstructed 3D points and substantially increase both the recall and precision of line segments. Incorporating additional wireframe constraints further improves performance by anchoring points to structural lines, thereby introducing stronger geometric consistency. We also evaluate our structural constraints on ETH3D~\cite{schops2017multi} using the official multi-view benchmark in \cref{tab:geometry_eth3d}. Our method improves both accuracy and completeness across all thresholds, demonstrating the effectiveness of group constraints in enhancing feature reconstruction by enforcing structural consistency.

\noindent
\textbf{Structure from Motion.}
As discussed in \cref{sec::discussions}, we integrate our bundle adjustment (BA) framework into a full incremental SfM pipeline and evaluate camera pose accuracy on four datasets: Hypersim \cite{roberts:2021}, ScanNet++ \cite{yeshwanth2023scannet++}, ETH3D \cite{schops2017multi}, and 7Scenes \cite{7scenes}. We use the first eight scenes of Hypersim as in \cite{liu2024robust} and adopt the splits from \cite{lin2025depth} for the remaining datasets. Our holistic bundle adjustment consistently improves pose accuracy over prior methods across all datasets, covering both indoor and outdoor scenes (\cref{tab:sfm_results}). The improvement is particularly pronounced on ScanNet++, which contains indoor environments with rich geometric structures such as parallel and coplanar surfaces.

\noindent
\textbf{Ablation study.}
We report results in \cref{tab:ablation_group_3dcost}, comparing our 2D group-induced reprojection error with direct 3D plane-distance constraints under different weights. The 3D formulation requires careful tuning: small weights underuse structural information, while large weights ($w{=}1000$) degrade pose accuracy on ScanNet++. In contrast, our 2D formulation requires no tuning and achieves the best or competitive results on both geometry (ETH3D) and pose (ScanNet++) metrics.

\begin{figure}[tb]
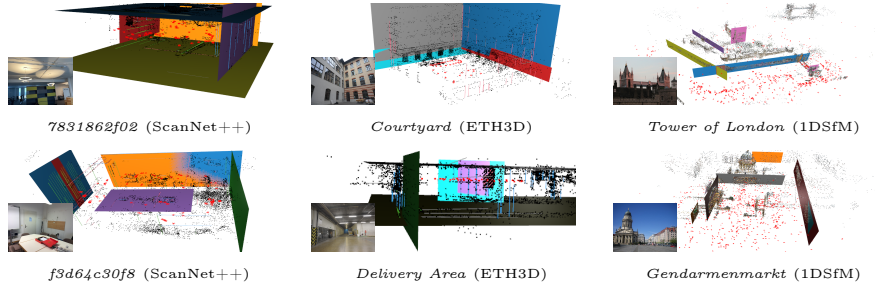

\centering
\def\visimg{figs/imgs/visualizations}
\def\rgbimg{figs/imgs/visualizations/rgb_images}
\newlength{\visrowh}\setlength{\visrowh}{38pt}
\newlength{\rgbrowh}\setlength{\rgbrowh}{18pt}
\newlength{\rgbroww}\setlength{\rgbroww}{24pt}
\setlength{\tabcolsep}{10pt}
\newcommand{\visoverlay}[4]{%
  \includegraphics[#1]{#2}%
  \llap{\smash{\raisebox{\dimexpr\visrowh-\rgbrowh-20pt\relax}{%
    \makebox[#4][l]{\hspace{-6pt}%
      \includegraphics[width=\rgbroww, height=\rgbrowh]{#3}}}}}%
}
\begin{tabular}{@{}ccc@{}}
\visoverlay{width=0.27\linewidth, height=\visrowh, trim={440 232 668 422}, clip}{\visimg/scannetpp_7831862f02}{\rgbimg/7813_000068}{0.27\linewidth} &
\visoverlay{width=0.27\linewidth, height=\visrowh, trim={506 250 551 382}, clip}{\visimg/eth3d_courtyard}{\rgbimg/courtyard_000025}{0.27\linewidth} &
\visoverlay{width=0.27\linewidth, height=\visrowh, trim={260 160 240 180}, clip}{\visimg/1dsfm_Tower_of_London_aligned}{\rgbimg/tower_501227104_73e266a54e_o}{0.27\linewidth} \\
\parbox{0.27\linewidth}{\centering\tiny \textit{7831862f02} (ScanNet++)} &
\parbox{0.27\linewidth}{\centering\tiny \textit{Courtyard} (ETH3D)} &
\parbox{0.27\linewidth}{\centering\tiny \textit{Tower of London} (1DSfM)} \\[3pt]
%
\visoverlay{width=0.27\linewidth, height=\visrowh, trim={537 197 580 321}, clip}{\visimg/scannetpp_f3d64c30f8}{\rgbimg/f3d6_000038}{0.27\linewidth} &
\visoverlay{width=0.27\linewidth, height=\visrowh, trim={597 423 880 481}, clip}{\visimg/eth3d_delivery_area_view1}{\rgbimg/delivery_area_000000}{0.27\linewidth} &
\visoverlay{width=0.27\linewidth, height=\visrowh, trim={550 190 450 210}, clip}{\visimg/1dsfm_Gendarmenmarkt_aligned}{\rgbimg/gendar_480352481_2892812c16_o}{0.27\linewidth} \\
\parbox{0.27\linewidth}{\centering\tiny \textit{f3d64c30f8} (ScanNet++)} &
\parbox{0.27\linewidth}{\centering\tiny \textit{Delivery Area} (ETH3D)} &
\parbox{0.27\linewidth}{\centering\tiny \textit{Gendarmenmarkt} (1DSfM)} \\
\end{tabular}
\caption{\textbf{Visualizations of bundle adjusted holistic 3D structures.} Parallel lines are colored the same. Note that lines and planes are optimized as infinite entities in the bundle adjustment, and their extents in the figures are only for illustration purpose.}
\label{fig::visualizations}
\end{figure}

\noindent
\textbf{Mapping Geometric Primitives.}
Beyond quantitative improvements, our framework produces semantically richer maps with explicit geometric primitives. Example holistic 3D reconstructions are shown in \cref{fig::visualizations}. While planes and vanishing points are the most common structural groups in real-world scenes, our formulation is general and supports other primitives. In \cref{fig::teaser}, we show bundle adjustment on a scene containing planes, spheres, and cylinders, where groups are detected via SAM3 \cite{carion2025sam} with semantic prompts (football$\to$sphere, can$\to$cylinder, laptop/postcard$\to$plane). This demonstrates its flexibility and shows the potential of combining geometric abstraction with modern visual-language models. 
We believe our framework opens up opportunities to further integrate geometric reasoning with visual-language understanding.
\section{Conclusion}
We present a unified bundle adjustment framework that integrates higher-order geometric structures while preserving efficiency. By placing group parameters alongside cameras and formulating all constraints as reprojection-based residuals, we maintain efficient Schur elimination and avoid the scale ambiguity and conditioning issues of 3D regularization. Experiments demonstrate comparable runtime with significantly richer reconstructions and improved accuracy.

\noindent
\textbf{Acknowledgements.}
Viktor Larsson was supported by ELLIIT and the Swedish Research Council (Grant No. 2023-05424). We sincerely thank Alexander Veicht, Paul-Edouard Sarlin, Eric Dexheimer, Xinyue Zhang, Linfei Pan, Elisabetta Fedele, Johannes Schönberger, Lixin Xue, and anonymous reviewers for their helpful feedbacks. 

\section*{Appendix}
\appendix
This supplementary material accompanies the main paper and is organized as follows.
We begin with implementation details for image description, matching, and structure-from-motion, as well as the synthetic benchmark setup in \cref{sec:supp_impl}.
We then present some example groups and their parameterizations in \cref{sec:supp_param}, followed by more details and discussions on the proposed bundle adjustment framework in \cref{sec:supp_discussions}.
Finally, we show some additional visualizations.

\section{Implementation Details}
\label{sec:supp_impl}

\subsection{Additional Details on Image Description and Matching}
\label{sec:supp_description}

\begin{figure}[tb]
\centering
\def\frontimg{figs/imgs/supp/frontend}
\setlength{\tabcolsep}{1pt}
\renewcommand{\arraystretch}{0}
\newcommand{\fw}{0.16\linewidth}
\begin{tabular}{@{}cccccc@{}}
\includegraphics[width=\fw]{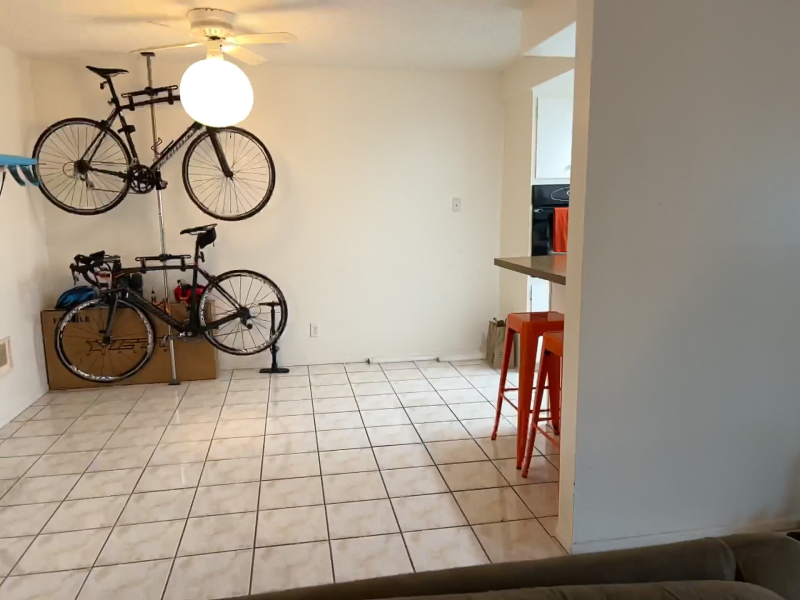} &
\includegraphics[width=\fw]{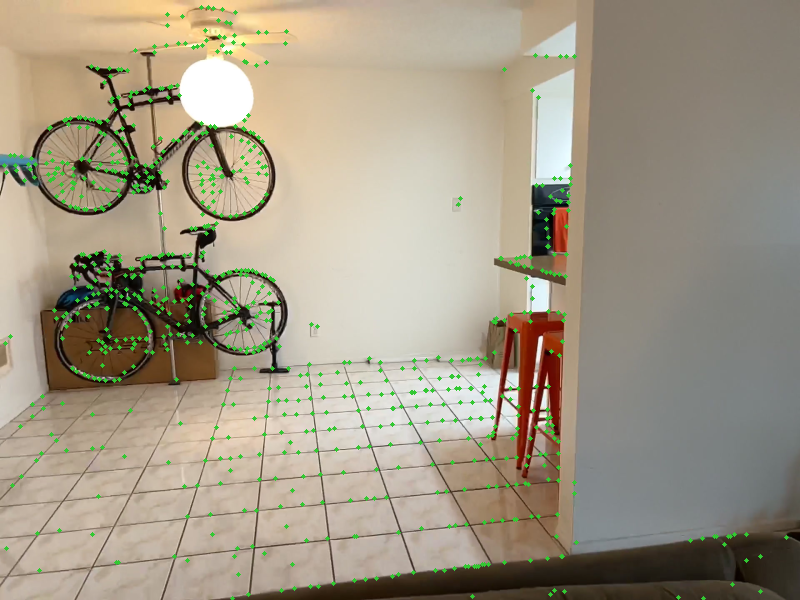} &
\includegraphics[width=\fw]{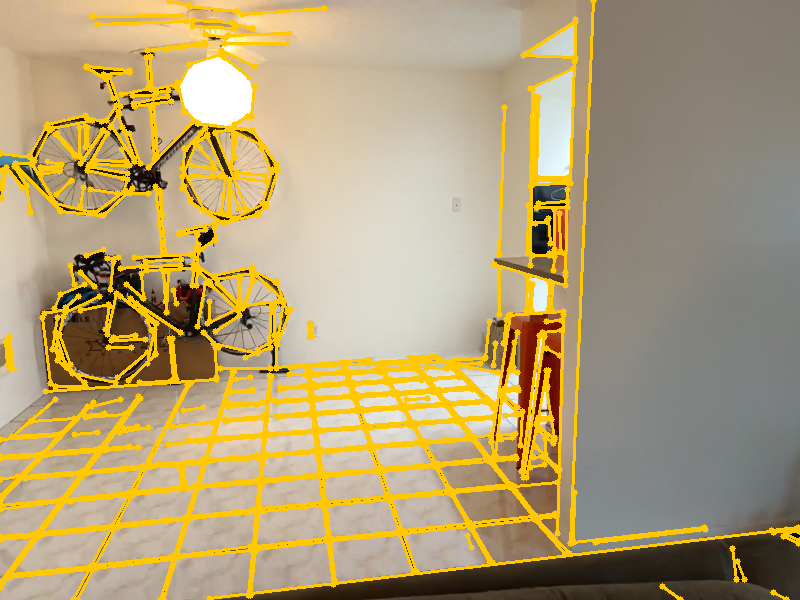} &
\includegraphics[width=\fw]{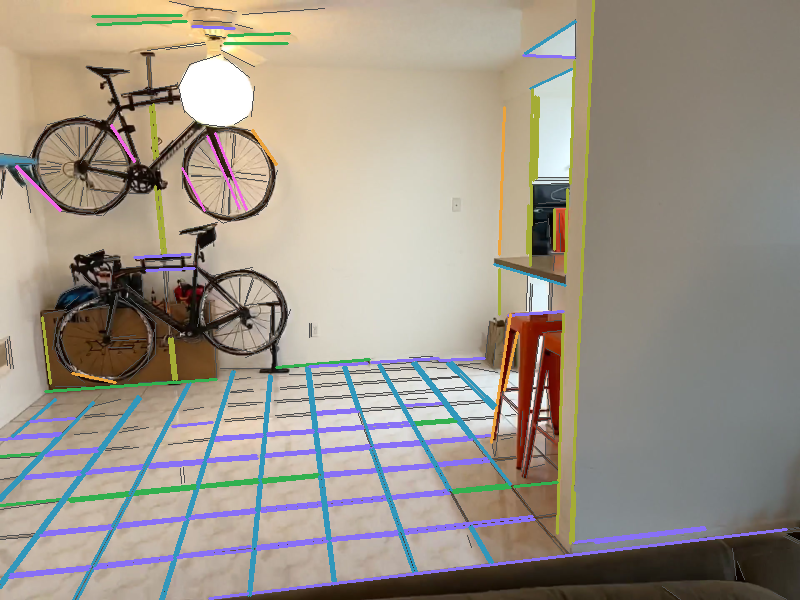} &
\includegraphics[width=\fw]{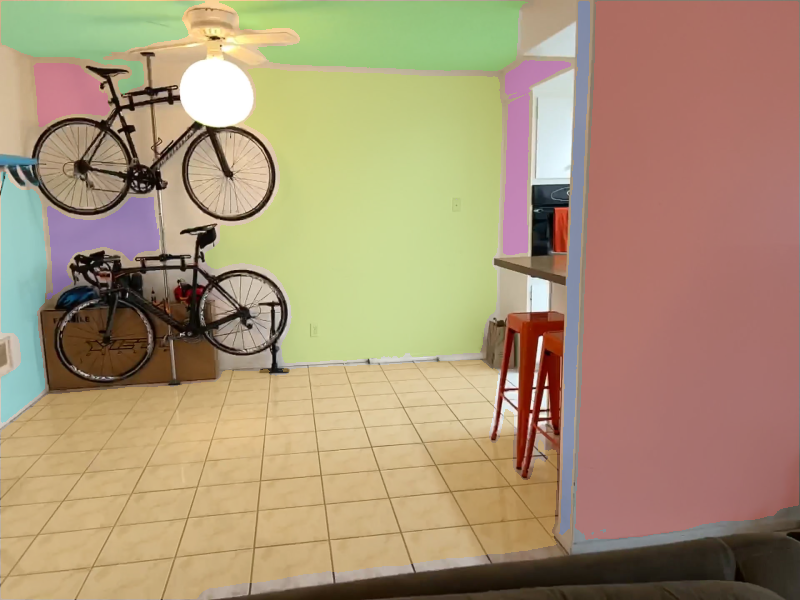} &
\includegraphics[width=\fw]{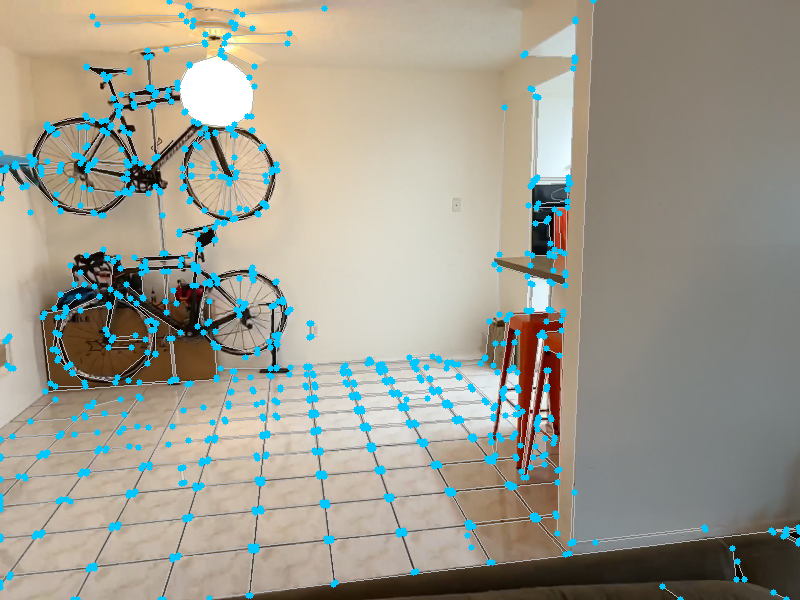} \\[1pt]
\includegraphics[width=\fw]{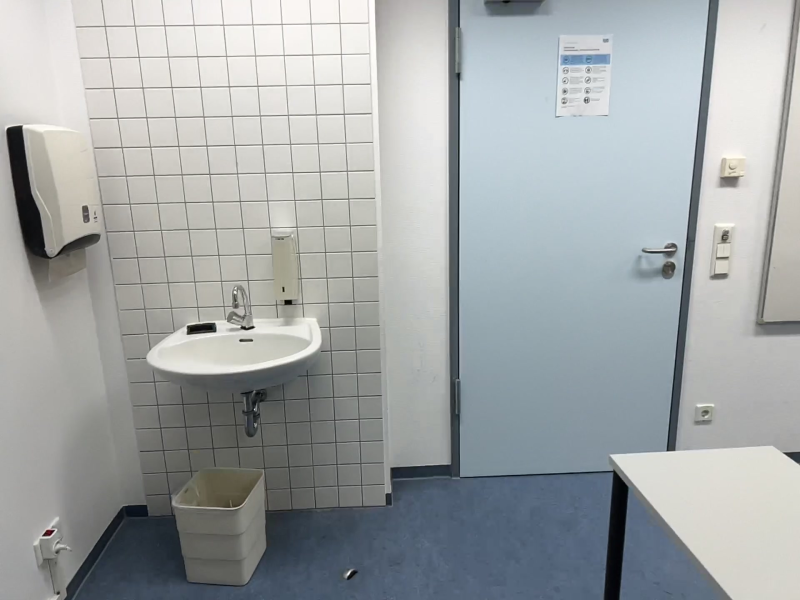} &
\includegraphics[width=\fw]{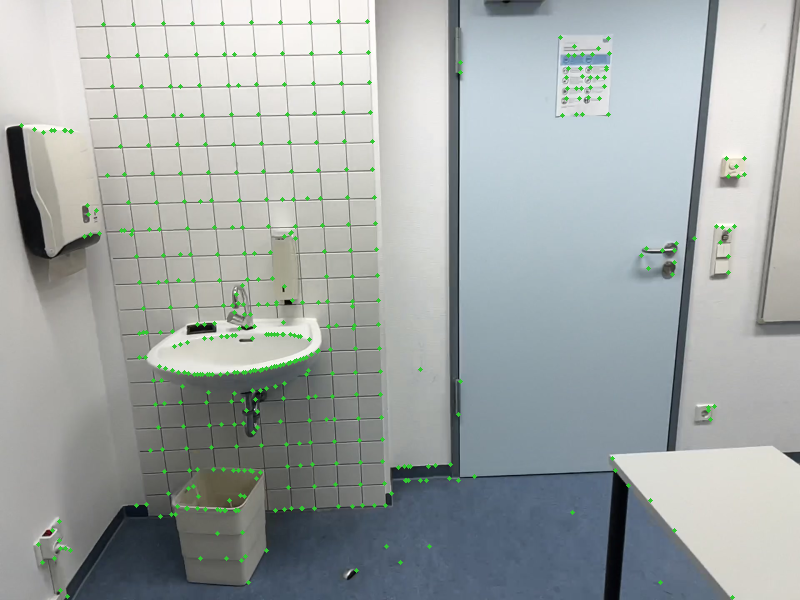} &
\includegraphics[width=\fw]{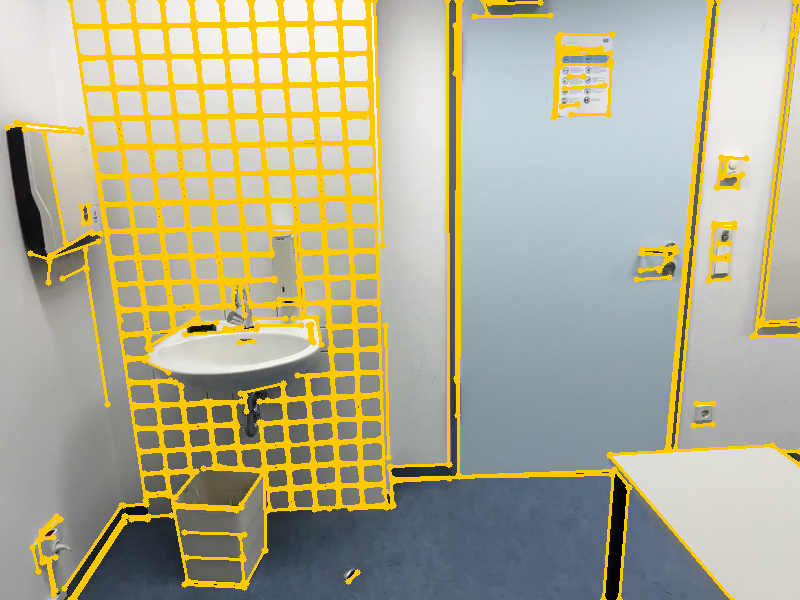} &
\includegraphics[width=\fw]{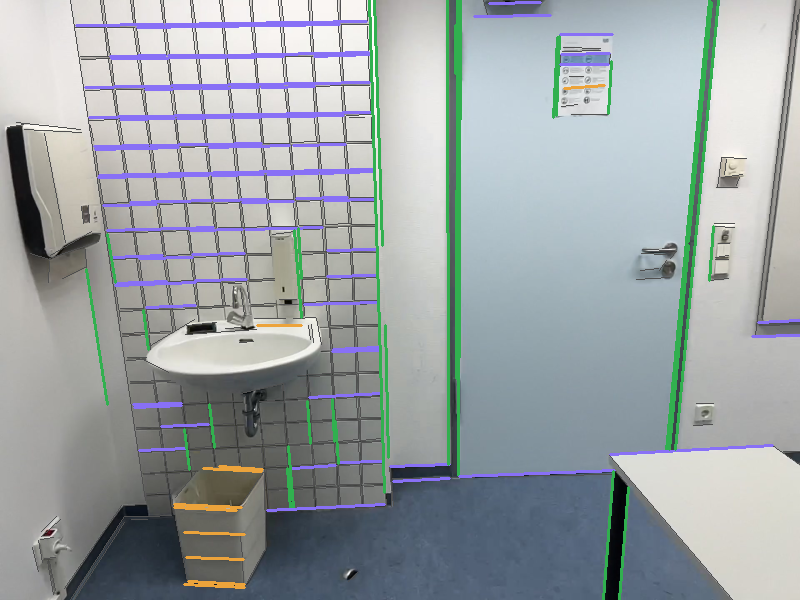} &
\includegraphics[width=\fw]{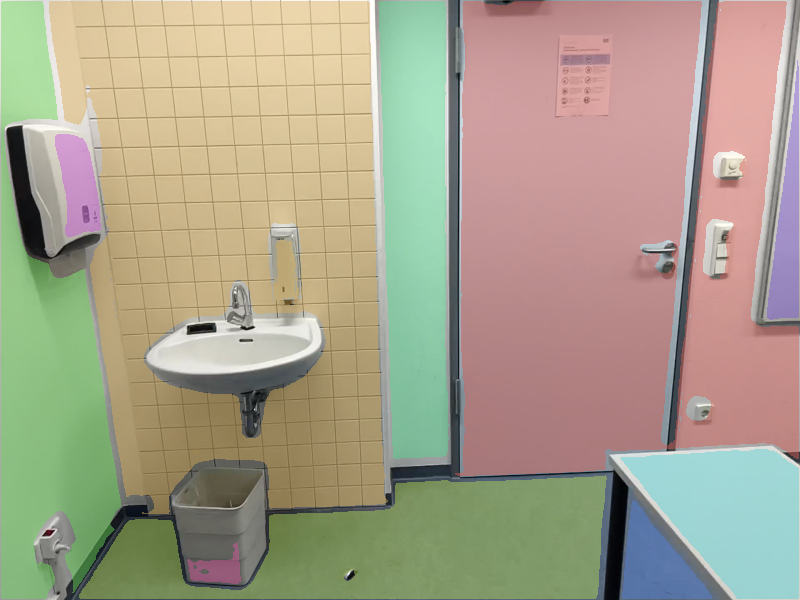} &
\includegraphics[width=\fw]{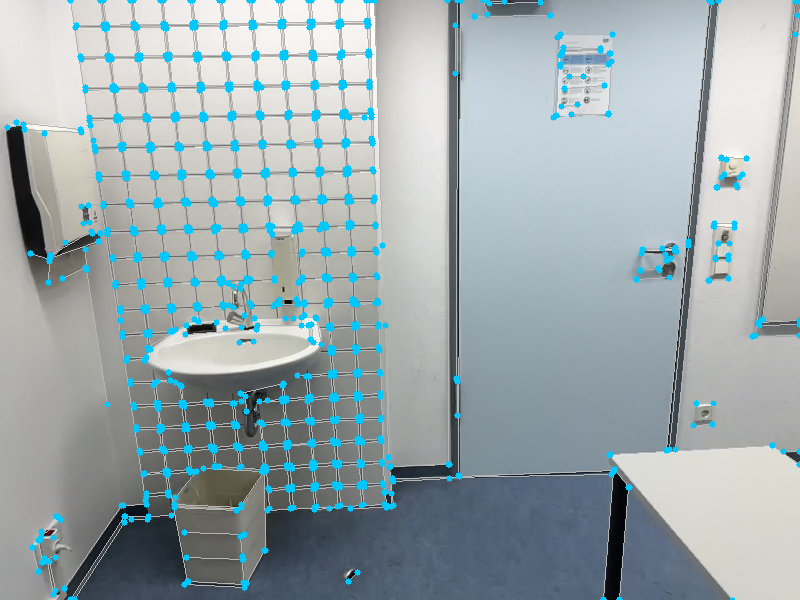} \\[1pt]
\includegraphics[width=\fw]{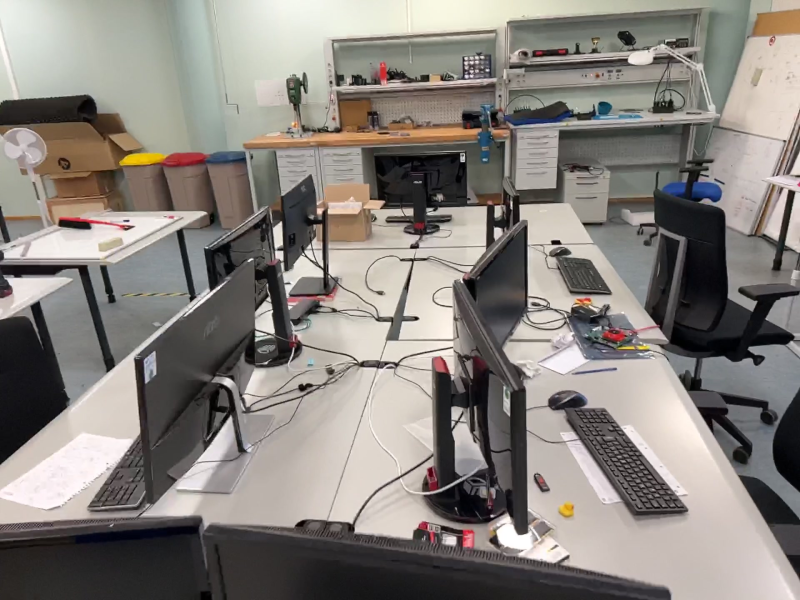} &
\includegraphics[width=\fw]{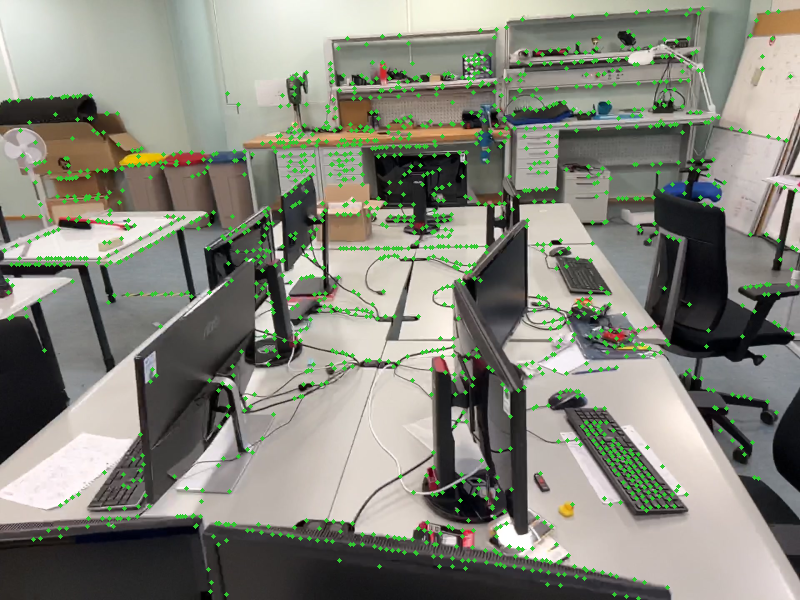} &
\includegraphics[width=\fw]{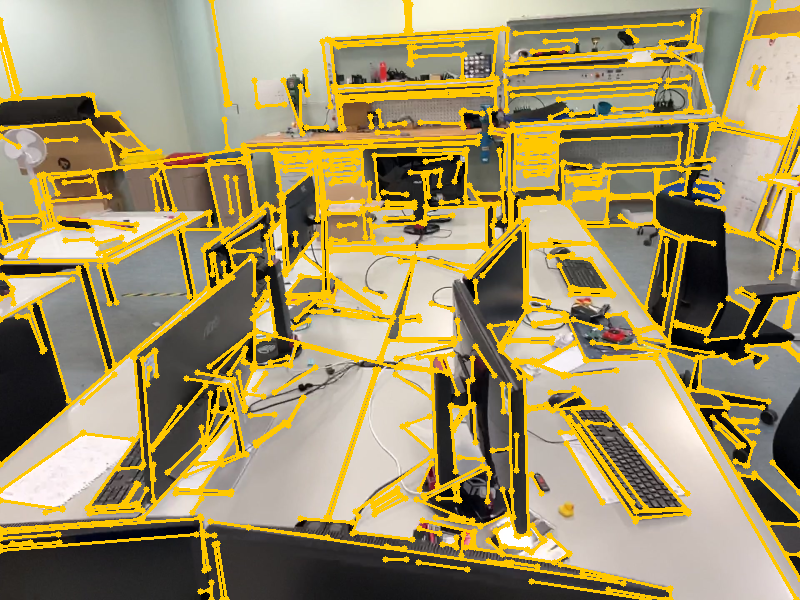} &
\includegraphics[width=\fw]{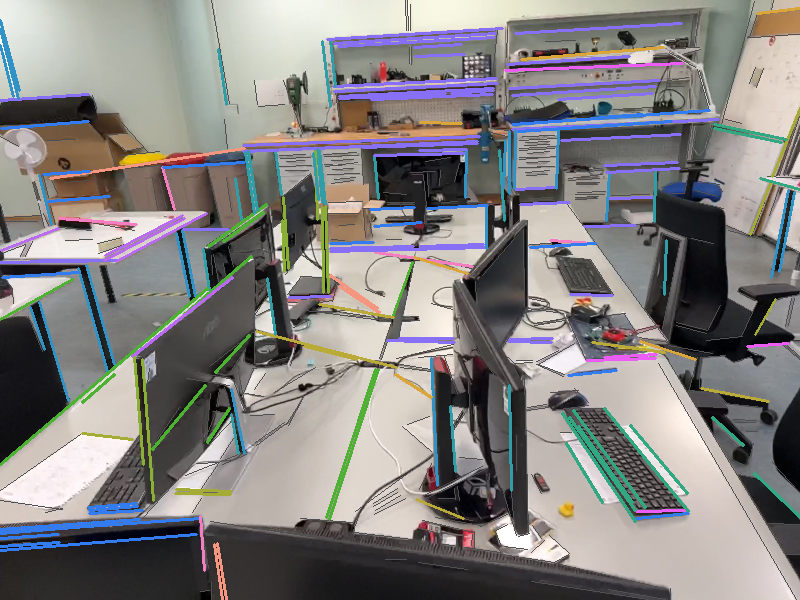} &
\includegraphics[width=\fw]{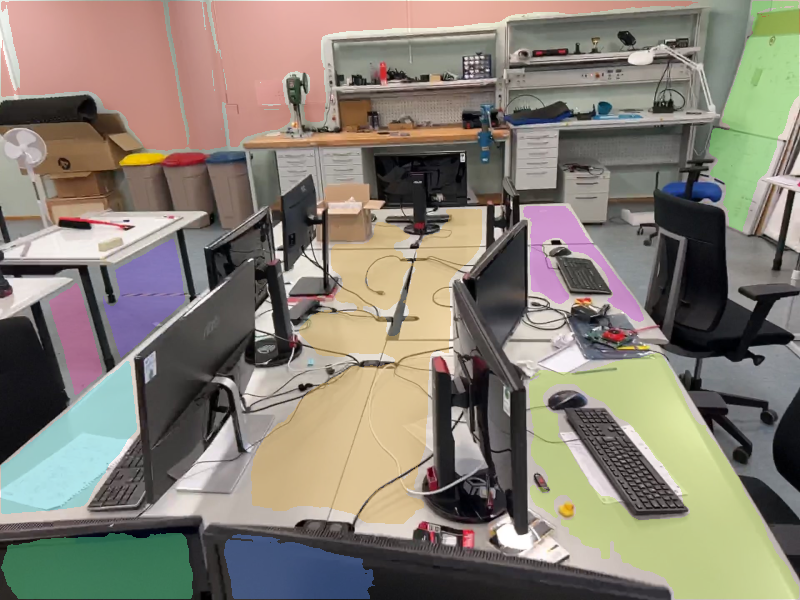} &
\includegraphics[width=\fw]{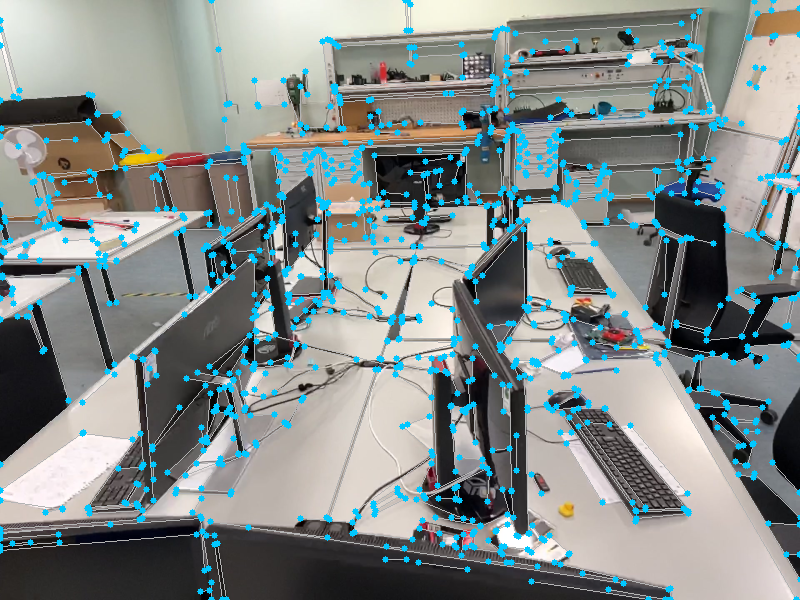} \\[2pt]
Input & Points & Lines & VPs & Planes & Junctions \\
\end{tabular}
\caption{\textbf{Example features, groups, and wireframes from image description.}}
\label{fig:detection}
\end{figure}

\noindent
\textbf{Feature Detection and Matching.}
We use ALIKED~\cite{zhao2023aliked} for point detection and LightGlue~\cite{lindenberger2023lightglue} for point matching.
For lines, we use DeepLSD~\cite{Pautrat_2023_DeepLSD} for detection and GlueStick~\cite{pautrat2023gluestick} for matching.
Vanishing points are detected with JLinkage~\cite{toldo2008robust}, and planes are segmented by fitting to predicted depth and surface normals from MoGe-2~\cite{wang2025moge,wang2025moge2}.
Wireframe junctions are derived from spatial incidences between detected points and lines within a distance threshold of 2 pixels. \Cref{fig:detection} shows example detection results, where the junctions are points associated with more than one line, connecting them into a wireframe structure.
Images are resized so that the longer side is at most 800 pixels.

\noindent
\textbf{Cross-Image Group Matching.}
Groups detected in different images must be matched to establish 3D constraints.
We support two options.
The first is voting via matched features: for each pair of groups in two images, we count the number of matched features associated with both groups and establish a match if this count exceeds 3.
The second is matching via dense correspondence: we warp the segmentation mask from one image to another using RoMa~\cite{edstedt2024roma}, and compute the overlap ratio between the warped mask and candidate group masks in the target image, matching groups with overlap ratio above 0.6 and intersection area above 1000 pixels.
Note that non-segmentation-based groups such as VPs can only use voting.
Empirically, voting yields higher matching accuracy due to its reliance on verified feature correspondences, while dense matching based warping offers the advantage of matching points, lines, and segmentation-based groups jointly.

\noindent
\textbf{Geometric Verification.}
When prior camera poses are available, we verify group associations through geometric consistency checks.
For VPs, we back-project 2D image directions into 3D and check angular consistency across views, rejecting matches with angular deviation above 10 degrees.
For planes, we compare monocular surface normal estimates with the triangulated plane normal, rejecting matches with angular deviation above 30 degrees.

\subsection{Additional Details on Structure-from-Motion}
\label{sec:supp_sfm}

\noindent
\textbf{Image Retrieval.}
For image pair selection, we use NetVLAD~\cite{arandjelovic2016netvlad} to retrieve the top 30 nearest neighbors for each image.

\noindent
\textbf{Incremental Mapping.}
We follow prior work on incremental triangulation of points~\cite{schonberger2016structure} and lines~\cite{Liu_2023_LIMAP,liu2024robust} and extend it to incremental group triangulation as new images are registered. For each newly registered image, point and line features are first triangulated following the standard incremental pipeline. Group triangulation is then performed based on the resulting triangulated features.
When a new 2D group observation is introduced, we count the number of shared triangulated features with existing 3D groups. If at least two features overlap, the observation is associated with the corresponding group.
If no match is found, the observation is temporarily stored until a consistent set of features emerges. Specifically, when at least three features consistently co-occur across unmatched observations, a new 3D group is initialized. Group parameters are estimated using LO-RANSAC~\cite{chum2003locally}. We adopt a relaxed 3D threshold at this stage and defer fine-grained group refinement to the subsequent optimization.
After new triangulations, we check for potential group merges. Two groups of the same type become merge candidates when their 2D observations are connected through feature correspondences across images. A merge is performed if the group parameters are geometrically compatible—for example, vanishing point (VP) directions must lie within an angular threshold—and if the merged parameters do not significantly increase the fitting error of either group's features. When a merge occurs, feature associations are combined and the group parameters are re-estimated from the aggregated 3D features.
Finally, we apply geometric consistency filtering. Groups are validated using a reprojection consistency check similar to the group-induced reprojection error. In addition, we apply a 3D distance filter based on three times the median absolute deviation (MAD). A group is retained only if it contains at least three inlier features and achieves an inlier ratio greater than 0.5.

\noindent
\textbf{Reliability-Aware Refinement.}
Following prior work~\cite{liu2024robust}, we classify feature tracks as reliable or unreliable based on their number of active observations and estimated uncertainty.
Refinement proceeds in two stages: a full optimization including cameras, reliable features, groups, and wireframe constraints, followed by a fixed-pose refinement of unreliable features.
This prevents weakly-constrained features from corrupting pose estimates while retaining them for future use as more observations accumulate.
We extend this mechanism to groups: group associations are updated after each stage, and groups with insufficient active support are excluded from the first stage.

\noindent
\textbf{Full Pipeline.}
Our incremental SfM pipeline is built on top of COLMAP~\cite{schonberger2016structure} and closely follows its principles, including next image selection, registration with structure-less fallback, triangulation, local bundle adjustment after each image, and periodic global bundle adjustment.
We extend the pipeline by integrating group triangulation and structure-aware bundle adjustment at each refinement stage.

\subsection{Details on Setup of the Synthetic Benchmark}
\label{sec:supp_synthetic}

We construct synthetic bundle adjustment problems to analyze scaling behavior in isolation.
Cameras are placed on a sphere surrounding a cubic scene (half-extent 3.0) with viewing directions pointing inward.
Within the cube, we generate Manhattan-aligned planes with normals along the $x$, $y$, or $z$ axes at random offsets.
Points and lines are distributed both on these planes and as free-floating features.
Wireframe junctions are created by sampling a regular grid of points on each plane and connecting them with lines along the grid rows and columns, forming point-line associations that model structural corners and edges.
The number of images varies from 100 to 2000, with feature counts scaling proportionally: $10N$ points, $N$ lines, and $N/5$ planes for $N$ images, plus $3N$ wireframe edges.
Each feature is observed by 4 to 8 cameras on average, with lines on the same plane sharing approximately 50\% of their views to model real co-visibility patterns.
We add Gaussian noise to 2D observations (1.0 pixel for points and line endpoints in 2D), initial camera poses (1.0 degree for rotations, 0.05 for translations), and initial 3D estimates (0.1 for points and line endpoints in 3D).
For fair runtime comparison, we run each BA configuration for exactly 20 iterations.

\section{Example Groups and Their Parameterizations}
\label{sec:supp_param}
\label{sec:supp_param}

We support various types of geometric groups that constrain 3D features (points and lines) to lie on specific surfaces or follow directional constraints. \Cref{tab:group_params} summarizes each group with an illustration, degrees of freedom, and the features it constrains. Positive quantities such as radii, semi-axes, and angles are optimized in log-space to ensure positivity during optimization. Below we describe the parameterization for each group type. Note that while \Cref{tab:group_params} lists the group types currently implemented, the framework is not limited to these and can be extended to support additional geometric primitives such as splines, parametric surfaces, object-level priors, and parameterized priors from 3D foundation models (e.g., monocular depths and normals).

\newcommand{\sketchscale}{0.38}
\newcommand{\featPts}{Points}
\newcommand{\featLns}{Lines}

\begin{table}[t]
\centering
\caption{\textbf{Supported group types with illustrations.} Positive quantities (radii, semi-axes, angles) are parameterized in log-space to ensure positivity.}
\label{tab:group_params}
\small
\setlength{\tabcolsep}{4pt}
\begin{tabular}{@{}c l c c p{4.2cm}@{}}
\toprule
\textbf{Sketch} & \textbf{Group} & \textbf{\#DoF} & \textbf{Features} & \textbf{Geometric Interpretation} \\
\midrule
\begin{tikzpicture}[scale=\sketchscale, baseline=-0.5ex]
    \coordinate (vp) at (1.6, 1.3);
    \draw[thick, ->, >=stealth, blue!70!black] (0, 0) -- (vp);
    \draw[gray, thick] (-0.4, 0.2) -- (1.0, 0.9);
    \draw[gray, thick] (-0.4, -0.2) -- (1.0, 0.4);
    \draw[gray, thick] (-0.2, -0.5) -- (0.8, 0.1);
    \fill[blue!70!black] (vp) circle (3pt);
    \node[above right, font=\tiny] at (vp) {$\mathbf{v}$};
\end{tikzpicture}
& VP & 2 & \featLns & Direction $\mathbf{v} \in \mathbb{S}^2$ \\[12pt]
\begin{tikzpicture}[scale=\sketchscale, baseline=-0.5ex]
    \fill[blue!10] (-0.7, -0.4) -- (1.3, -0.2) -- (1.5, 0.8) -- (-0.5, 0.6) -- cycle;
    \draw[thick, blue!50!black] (-0.7, -0.4) -- (1.3, -0.2) -- (1.5, 0.8) -- (-0.5, 0.6) -- cycle;
    \draw[thick, ->, >=stealth, red!70!black] (0.4, 0.2) -- (0.4, 1.3);
    \node[right, font=\tiny, red!70!black] at (0.4, 1.1) {$\mathbf{n}$};
    \fill[black] (0.1, 0.05) circle (2pt);
    \fill[black] (0.8, 0.35) circle (2pt);
    \draw[thick, orange] (-0.2, 0.45) -- (1.0, 0.0);
\end{tikzpicture}
& Plane & 3 & \featPts, \featLns & Normal $\mathbf{n} \in \mathbb{S}^2$, offset $w$ \\[12pt]
\begin{tikzpicture}[scale=\sketchscale, baseline=-0.5ex]
    \draw[thick, blue!50!black] (0,0) circle (0.9);
    \draw[dashed, blue!30] (0,0) ellipse (0.9 and 0.27);
    \fill[red!70!black] (0,0) circle (2.5pt);
    \node[below left, font=\tiny, red!70!black] at (0,0) {$\mathbf{c}$};
    \draw[<->, >=stealth, thick] (0,0) -- (0.636, 0.636);
    \node[above right, font=\tiny] at (0.25, 0.25) {$r$};
    \fill[black] (0.636, 0.636) circle (2pt);
    \fill[black] (-0.78, 0.45) circle (2pt);
\end{tikzpicture}
& Sphere & 4 & \featPts & Center $\mathbf{c} \in \mathbb{R}^3$, radius $r > 0$ \\[12pt]
\begin{tikzpicture}[scale=\sketchscale, baseline=-0.5ex]
    \draw[thick, blue!50!black] (-0.55, -0.9) -- (-0.55, 0.9);
    \draw[thick, blue!50!black] (0.55, -0.9) -- (0.55, 0.9);
    \draw[dashed, blue!50!black] (0, 0.9) ellipse (0.55 and 0.18);
    \draw[dashed, blue!50!black] (0, -0.9) ellipse (0.55 and 0.18);
    \draw[thick, ->, >=stealth, red!70!black] (0, -1.1) -- (0, 1.3);
    \node[right, font=\tiny, red!70!black] at (0, 1.1) {$\mathbf{d}$};
    \draw[<->, >=stealth, thick] (0, 0) -- (0.55, 0);
    \node[above, font=\tiny] at (0.27, 0) {$r$};
    \fill[black] (0.55, 0.4) circle (2pt);
    \fill[black] (-0.55, -0.25) circle (2pt);
\end{tikzpicture}
& Cylinder & 5 & \featPts & Axis ($SO(3) \times SO(2)$), radius $r > 0$ \\[12pt]
\begin{tikzpicture}[scale=\sketchscale, baseline=-0.5ex]
    \coordinate (apex) at (0, 1.1);
    \draw[thick, blue!50!black] (apex) -- (-0.8, -0.8);
    \draw[thick, blue!50!black] (apex) -- (0.8, -0.8);
    \draw[dashed, blue!50!black] (0, -0.8) ellipse (0.8 and 0.22);
    \fill[red!70!black] (apex) circle (2.5pt);
    \node[above, font=\tiny, red!70!black] at (apex) {$\mathbf{a}$};
    \draw[thick, ->, >=stealth, red!70!black] (apex) -- (0, -1.1);
    \draw[dashed, gray] (apex) -- (0, -0.8);
    \draw[thick, gray] (0, 0.5) arc (270:292:0.6);
    \node[right, font=\tiny, gray] at (0.15, 0.35) {$\alpha$};
    \fill[black] (0.4, -0.2) circle (2pt);
\end{tikzpicture}
& Cone & 6 & \featPts & Apex $\mathbf{a} \in \mathbb{R}^3$, axis $\mathbf{d} \in \mathbb{S}^2$, half-angle $\alpha$ \\[12pt]
\begin{tikzpicture}[scale=\sketchscale, baseline=-0.5ex]
    \draw[thick, blue!50!black] (0,0) ellipse (1.2 and 0.65);
    \draw[dashed, blue!30] (0,0) ellipse (0.4 and 0.65);
    \fill[red!70!black] (0,0) circle (2.5pt);
    \draw[<->, >=stealth, thick] (0,0) -- (1.2, 0);
    \node[below, font=\tiny] at (0.6, 0) {$s_x$};
    \draw[<->, >=stealth, thick] (0,0) -- (0, 0.65);
    \node[left, font=\tiny] at (0, 0.32) {$s_y$};
    \draw[thick, ->, >=stealth, gray] (1.4, 0.4) arc (30:85:0.35);
    \node[right, font=\tiny, gray] at (1.35, 0.55) {$R$};
\end{tikzpicture}
& Ellipsoid & 9 & \featPts & $R \in SO(3)$, center $\mathbf{c} \in \mathbb{R}^3$, semi-axes $s_x, s_y, s_z > 0$ \\[12pt]
\begin{tikzpicture}[scale=\sketchscale, baseline=-0.5ex]
    \fill[blue!5] (-0.6, -0.45) -- (0.6, -0.45) -- (0.6, 0.6) -- (-0.6, 0.6) -- cycle;
    \draw[thick, blue!50!black] (-0.6, -0.45) -- (0.6, -0.45) -- (0.6, 0.6) -- (-0.6, 0.6) -- cycle;
    \draw[thick, blue!50!black] (-0.6, -0.45) -- (-1.0, -0.75);
    \draw[thick, blue!50!black] (0.6, -0.45) -- (0.2, -0.75);
    \draw[thick, blue!50!black] (0.6, 0.6) -- (0.2, 0.3);
    \draw[thick, blue!50!black] (-0.6, 0.6) -- (-1.0, 0.3);
    \fill[blue!15] (-1.0, -0.75) -- (0.2, -0.75) -- (0.2, 0.3) -- (-1.0, 0.3) -- cycle;
    \draw[thick, blue!50!black] (-1.0, -0.75) -- (0.2, -0.75) -- (0.2, 0.3) -- (-1.0, 0.3) -- cycle;
    \draw[thick, ->, >=stealth, gray] (1.1, 0.5) arc (30:85:0.35);
    \node[right, font=\tiny, gray] at (1.05, 0.65) {$R$};
    \draw[thick, orange] (-0.85, -0.6) -- (0.05, -0.15);
    \fill[black] (-0.6, 0.05) circle (2pt);
\end{tikzpicture}
& Cuboid & 9 & \featPts, \featLns & $R \in SO(3)$, 6 face offsets \\
\bottomrule
\end{tabular}
\end{table}

\noindent
\textbf{Vanishing Point (VP).}
A VP is a point at infinity representing a 3D direction, parameterized as $\mathbf{v} \in \mathbb{S}^2$.
The residual for a line with direction $\mathbf{d}$ measures angular deviation:
\begin{equation}
e = \|\mathbf{d} \times \mathbf{v}\|.
\end{equation}
This is equivalent to $e = \|\mathbf{v} - \text{Proj}_{\mathbf{d}}(\mathbf{v})\|$ where $\text{Proj}_{\mathbf{d}}(\mathbf{v}) = (\mathbf{v}^\top \mathbf{d})\mathbf{d}$, \ie the norm of the component of $\mathbf{v}$ orthogonal to $\mathbf{d}$.
Geometrically, both measure $\sin\theta$ where $\theta$ is the angle between the two unit vectors.

\noindent
\textbf{Plane.}
A plane is parameterized as $(\mathbf{n}, w)$ where $\mathbf{n} \in \mathbb{S}^2$ is the normal and $w \in \mathbb{R}$ is the signed distance to the origin.
The plane equation is $\mathbf{n}^\top \mathbf{x} + w = 0$.
Points are projected via:
\begin{equation}
\mathbf{p}_\text{proj} = \mathbf{p} - (\mathbf{n}^\top \mathbf{p} + w)\mathbf{n}.
\end{equation}
For lines in Pl\"ucker coordinates $(\mathbf{d}, \mathbf{m})$, the projected direction removes the normal component:
\begin{equation}
\mathbf{d}' = \mathbf{d} - (\mathbf{d} \cdot \mathbf{n})\mathbf{n}.
\end{equation}
The projected moment is:
\begin{equation}
\mathbf{m}' = \left(1 - \frac{\alpha^2}{\|\mathbf{d}\|^2}\right)\mathbf{m} + \frac{\alpha\gamma}{\|\mathbf{d}\|^2}\mathbf{d} + \left(w - \frac{\delta}{\|\mathbf{d}\|^2}\right)(\mathbf{d} \times \mathbf{n}),
\end{equation}
where $\alpha = \mathbf{d} \cdot \mathbf{n}$, $\gamma = \mathbf{m} \cdot \mathbf{n}$, and $\delta = (\mathbf{d} \times \mathbf{n}) \cdot \mathbf{m}$.

\noindent
\textbf{Sphere.}
A sphere is parameterized by its center $\mathbf{c} \in \mathbb{R}^3$ and radius $r > 0$.
Points are projected radially:
\begin{equation}
\mathbf{p}_\text{proj} = \mathbf{c} + r \cdot \frac{\mathbf{p} - \mathbf{c}}{\|\mathbf{p} - \mathbf{c}\|}.
\end{equation}

\noindent
\textbf{Cylinder.}
A cylinder is defined by an infinite axis (parameterized via $SO(3) \times SO(2)$) and radius $r > 0$.
The axis is represented using minimal Pl\"ucker coordinates~\cite{bartoli2005structure}.
Points are projected radially onto the cylinder surface perpendicular to the axis.

\noindent
\textbf{Cone.}
A cone is parameterized by its apex $\mathbf{a} \in \mathbb{R}^3$, axis direction $\mathbf{d} \in \mathbb{S}^2$, and half-angle $\alpha \in (0, \frac{\pi}{2})$.
Points are projected onto the cone surface along the direction perpendicular to the cone's generatrix.

\noindent
\textbf{Ellipsoid.}
An ellipsoid is parameterized by its pose (rotation $R \in SO(3)$ and center $\mathbf{c}$) and semi-axes $(s_x, s_y, s_z)$ with $s_i > 0$.
The surface equation in the local frame is:
\begin{equation}
\mathbf{x}^\top \text{diag}(s_x^{-2}, s_y^{-2}, s_z^{-2}) \mathbf{x} = 1.
\end{equation}
Points are projected via scaled-radial projection: the point is transformed to the local frame, scaled to a unit sphere, normalized, scaled back, and transformed to the world frame.

\noindent
\textbf{Cuboid.}
A cuboid is parameterized by a rotation $R \in SO(3)$ and six face offsets $d_0, \ldots, d_5$ \cite{hanning2025pixcuboid}.
The rotation defines three orthogonal face normal directions $\pm R_{:,0}, \pm R_{:,1}, \pm R_{:,2}$, and each face $i$ is a plane with normal $\mathbf{n}_i$ and offset $d_i$.
For point projection, we compute the signed distance to each of the six faces and project the point onto the nearest face.
For line projection, we first find the closest point on the infinite line to the cuboid center, then determine which face is nearest to this representative point, and finally project the line onto that face using plane projection.

\begin{table}[tb]
\centering
\caption{\textbf{Schur complement density analysis.} $D$ denotes the fraction of camera pairs with non-zero blocks. Wireframe constraints (WF) only add modest overhead.}
\label{tab:schur_density}
\small
\setlength{\tabcolsep}{5pt}
\begin{tabular}{@{}lrrrrccc@{}}
\toprule
Scene & \#img & \#pts & \#lns & \#wf & $D$ w/o.\ WF & $D$ w/.\ WF & $\Delta$\% \\
\midrule
Gendarmenmarkt & 951 & 88k & 3.7k & 12.0k & 41.1\% & 42.0\% & 2.2 \\
Madrid Metropolis & 750 & 65k & 2.1k & 6.0k & 23.8\% & 24.3\% & 1.9 \\
Tower of London & 963 & 99k & 2.8k & 9.1k & 12.3\% & 13.2\% & 7.4 \\
\midrule
Average & & & & & 25.7\% & 26.5\% & 3.8 \\
\bottomrule
\end{tabular}
\end{table}

\begin{figure*}[tb]
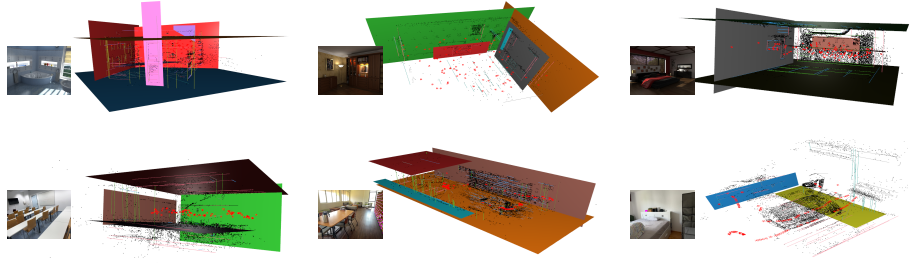

\centering
\def\visimg{figs/imgs/supp/visualizations}
\def\rgbimg{figs/imgs/supp/visualizations/rgb_images}
\newlength{\svisrowh}\setlength{\svisrowh}{48pt}
\newlength{\srgbrowh}\setlength{\srgbrowh}{18pt}
\newlength{\srgbroww}\setlength{\srgbroww}{24pt}
\setlength{\tabcolsep}{10pt}
\newcommand{\svisoverlay}[4]{%
  \includegraphics[#1]{#2}%
  \llap{\smash{\raisebox{\dimexpr\svisrowh-\srgbrowh-20pt\relax}{%
    \makebox[#4][l]{\hspace{-6pt}%
      \includegraphics[width=\srgbroww, height=\srgbrowh]{#3}}}}}%
}
\begin{tabular}{@{}ccc@{}}
\svisoverlay{width=0.28\linewidth, height=\svisrowh, trim={456 207 392 125}, clip}{\visimg/hypersim_ai_001_001}{\rgbimg/hypersim_01_0019.color}{0.30\linewidth} &
\svisoverlay{width=0.28\linewidth, height=\svisrowh, trim={514 140 386 186}, clip}{\visimg/hypersim_ai_001_004}{\rgbimg/hypersim_04_0000.color}{0.30\linewidth} &
\svisoverlay{width=0.28\linewidth, height=\svisrowh, trim={439 280 569 375}, clip}{\visimg/hypersim_ai_001_006}{\rgbimg/hypersim_06_0000.color}{0.30\linewidth} \\[3pt]
%
\svisoverlay{width=0.28\linewidth, height=\svisrowh, trim={300 365 483 236}, clip}{\visimg/scannetpp_21d970d8de}{\rgbimg/scannetpp_21d9_000030}{0.30\linewidth} &
\svisoverlay{width=0.28\linewidth, height=\svisrowh, trim={578 157 491 317}, clip}{\visimg/scannetpp_9071e139d9}{\rgbimg/scannetpp_9071_000084}{0.30\linewidth} &
\svisoverlay{width=0.28\linewidth, height=\svisrowh, trim={428 143 548 163}, clip}{\visimg/scannetpp_bcd2436daf}{\rgbimg/scannetpp_bcd2_000012}{0.30\linewidth} \\
%
\end{tabular}
\caption{\textbf{Additional visualizations of bundle adjusted holistic 3D structures} on Hypersim~\cite{roberts:2021} (top) and ScanNet++~\cite{yeshwanth2023scannet++} (bottom). Parallel lines are colored the same. Note that lines and planes are optimized as infinite entities in the bundle adjustment, and their extents in the figures are only for illustration purpose.}
\label{fig::supp_visualizations}
\end{figure*}

\section{More Details and Discussions}
\label{sec:supp_discussions}
\label{sec:supp_discussions}

\noindent
\textbf{Balancing Different Types of Costs.}
As discussed in the main paper, 3D group costs (e.g., point-to-plane distance) and wireframe costs (e.g., 3D point-to-line distance) are problematic because their magnitude depends on the global scene scale, and varies across regions depending on viewing geometry.
In contrast, the angular cost for VP constraints, $e = \|\mathbf{d} \times \mathbf{v}\|$, is scale-free: it measures the sine of the angle between unit vectors and does not depend on the global scale of the reconstruction.
This makes angular residuals well-suited for directional constraints without the calibration difficulties of 3D metric residuals.
However, angular residuals are dimensionless and not directly comparable to pixel-based reprojection errors.
We therefore exclude angular terms from the termination criteria and monitor convergence based on reprojection residuals only, as described in the main paper.
For optimization, we apply a weight of $5 \times 10^3$ to VP angular constraints and use Cauchy loss with scale 0.2.
For reprojection-based group and wireframe costs (measured in pixels), we use weights of 0.5 and 0.1 respectively, relative to the base point and line reprojection weights of 1.0.

\noindent
\textbf{Relation to Manhattan/Atlanta World Assumption.}
The Manhattan world assumption~\cite{coughlan2000manhattan} posits three mutually orthogonal dominant directions, while the Atlanta world~\cite{schindler2004atlanta} relaxes this to a vertical direction with multiple horizontal directions.
Our framework implicitly enforces these assumptions through the combination of feature-group and inter-group constraints.
The VP-line constraints align lines with their associated vanishing points, while inter-group orthogonality constraints enforce perpendicularity between VP pairs.
Together, these naturally encode the Manhattan or Atlanta structure without requiring explicit world assumption handling.

\noindent
\textbf{Effects of Cross-Feature Constraints on the Schur Complement $\mathbf{S}$.}
As noted in the main paper, cross-feature (wireframe) constraints extend the effective visibility graph, making the Schur complement $\mathbf{S}$ denser in camera-camera blocks.
However, for a wireframe edge, new correlations between cameras $C_i$ and $C_j$ only arise when the two cameras share no co-observed features, and the associated point and line each appear in exactly one of them.
In practice, this additional density is modest.
\Cref{tab:schur_density} analyzes this on three scenes from the 1DSfM dataset, showing that wireframe constraints increase the Schur complement density by only 2--7\% relative to the baseline.

\noindent
\textbf{More Qualitative Results.}
We show some additional visualizations of the bundle adjusted 3D structures in \cref{fig::supp_visualizations}.

%
%
\bibliographystyle{splncs04}
\bibliography{main}
\end{document}